\RequirePackage{fix-cm}

\documentclass[twocolumn]{svjour3}          
\smartqed 
\usepackage{graphicx}
\usepackage{mathptmx}      
\DeclareMathAlphabet{\mathcal}{OMS}{cmsy}{m}{n}
\usepackage{latexsym}
\usepackage{microtype}
\usepackage{mathtools}
\usepackage{amssymb}
\usepackage{cases} 
\usepackage[caption=off]{subfig}
\usepackage{theoremref}
\usepackage[title]{appendix}
\usepackage{paralist}
\usepackage[normalem]{ulem}
\usepackage{hyperref}
\hypersetup{colorlinks=true,linkcolor=blue,citecolor=blue,urlcolor=black}
\usepackage{hyphenat}
\usepackage[misc]{ifsym}
\usepackage[dont-mess-around]{fnpct}
\usepackage{etoolbox}

\newcommand*{\affmark}[1][*]{\textsuperscript{#1}}
\newcommand{\orcid}[1]{\href{https://orcid.org/#1}{\includegraphics[width=9pt]{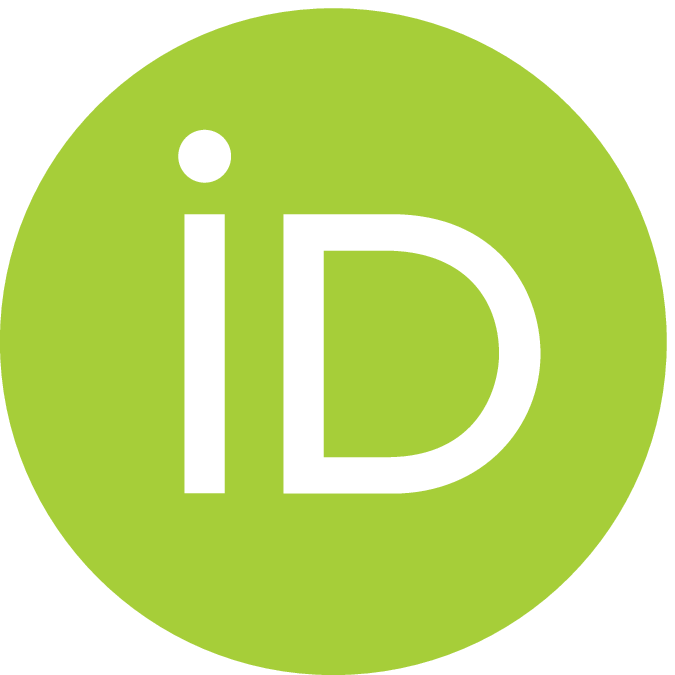}}}
\newcommand{\ie}{\textit{i}.\textit{e}.}
\newcommand{\eg}{\textit{e}.\textit{g}.}
\spnewtheorem{assumption}{Assumption}{\it}{\rm}
\spnewtheorem*{definition*}{Definition}{\bf}{\rm}
\spnewtheorem*{notation}{Notation}{\bf}{\rm}
\spnewtheorem*{theorem*}{Theorem}{\bf}{\it}

\everymath{\displaystyle}

\begin{document}
\tolerance 1414
\hbadness 1414
\emergencystretch 1.5em

\title{Footstep Adjustment for Biped Push Recovery on Slippery Surfaces}

\author{
Erfan Ghorbani\protect\affmark[1]\protect\orcid{0000-0003-1134-5031}\and 
Hossein Karimpour\protect\affmark[2]\protect\orcid{0000-0002-4152-2070}\and 
Venus Pasandi\protect\affmark[1]\and 
Mehdi Keshmiri\protect\affmark[1]
}

\authorrunning{
Erfan Ghorbani\and
}

\institute{
\begin{tabular}{l@{\hskip 5pt} l}
\Letter & Erfan Ghorbani\\
 & erfan.ghorbani@hotmail.com \vspace{5pt}\\
 & Hossein Karimpour \\
 & h.karimpour@eng.ui.ac.ir \vspace{5pt}\\
 & Venus Pasandi \\
 & venus.pasandi@me.iut.ac.ir \vspace{5pt}\\\
 & Mehdi Keshmiri \\
 & mehdik@cc.iut.ac.ir \vspace{5pt}\\
\affmark[1] & Department of Mechanical Engineering, Isfahan University  \\
                  & of Technology, Isfahan 84156-83111, Iran \vspace{5pt} \\
\affmark[2] & Department of Mechanical Engineering, University of Isfahan,\\
                  & Isfahan 81746-73441, Iran
\end{tabular}
}

\date{}
\maketitle

\begin{abstract}

Despite extensive studies on motion stabilization of bipeds, they still suffer from the lack of disturbance coping capability on slippery surfaces. 
In this paper, a novel controller for  stabilizing a bipedal motion in its sagittal plane is developed with regard to the surface friction limitations.
By taking into account the physical limitation of the surface in the stabilization trend, a more advanced level of reliability is achieved that provides higher functionalities such as push recovery on low-friction surfaces and prevents the stabilizer from overreacting.
The discrete event-based strategy consists of modifying the step length and time period at the beginning of each footstep in order to reestablish stability necessary conditions while taking into account the surface friction limitation as a constraint  to prevent slippage.
Adjusting footsteps to prevent slippage in confronting external disturbances is perceived as a novel strategy for keeping stability, quite similar to human reaction.
The developed methodology consists of rough closed-form solutions utilizing elementary math operations for obtaining the control inputs, allowing to reach a balance between convergence and computational cost, which is quite suitable for real-time operations even with modest computational hardware.
Several numerical simulations, including push recovery and switching between different gates on low-friction surfaces, are performed to demonstrate the effectiveness of the proposed controller.
In correlation with human-gait experience, the results also reveal some physical aspects favoring stability and the fact of switching between gaits to reduce the risk of falling in confronting different conditions.

\end{abstract}


\section{Introduction}
\label{intro}

Biped robots can adapt to the urban environment of modern societies due to their particular locomotion mode that is similar to the human gait, and therefore, are good candidates for collaborating with humans by walking alongside them. However, motion planning while maintaining stability is a challenging issue due to the switching nature of walking, besides its highly nonlinear and multi-dimensional dynamics. Moreover, in real conditions, factors such as low-value friction coefficient or surface irregularities increase the likelihood of slippery and stumbling. The dynamic stabilization problem of walking becomes thus even more complicated.
In a general framework, the stability concept for bipeds is defined as fall prevention and is expressed through the viability theory \cite{wieber2002stability}. With $\mathcal{F}$ as the set of states which leads to immediate fall, a state is viable if and only if the biped can realize a movement starting from this state that never gets inside the set $\mathcal{F}$. The union of all viable states is called the viability kernel. Accordingly, stabilization is interpreted as defining a control law to make the viability kernel completely invariant.
However, the mathematical formulation of the viability kernel is generally impossible  due to the complexities of bipeds dynamics, so the present methods utilize only subsets of the viability kernel. \emph{Postural stability} and \emph{cyclic stability} are among the well-known stabilization methods, each one dealing with a different subset of the viability kernel.

The general trend of postural stability relies on ensuring instantaneous stability through applying sufficient restrictions to the biped configuration at each moment in order to ensure it will not fall.
In contrast, cyclic stability methods deal with the overall stability of motion rather than satisfying it continuously. In this way, walking is generated by following a given limit cycle. Accordingly, the method is formally known as \emph{Limit Cycle Walking} and is defined as ''a nominally periodic sequence of steps that is stable as a whole but not locally stable at every instant in time'' \cite{hobbelen2007limit}. The limit cycles in bipedal locomotion correspond to the sequence of a swinging phase followed by a step impact and a leg switching transition, where the latter yields an instantaneous transfer of state in the position-coordinate direction and the former results in an instantaneous jump in the velocity-coordinate direction \cite{hurmuzlu1986role}. 
Poincar\'e map \cite{mcgeer1990passive}, hybrid zero dynamics \cite{westervelt2003hybrid}, the notion of symmetry \cite{razavi2017symmetry}, and coupled oscillators \cite{aoi2005locomotion} are among the techniques used for generating cyclic gaits.
On the other hand, static stability \cite{mcghee1968stability}, ZMP \cite{vukobratovic2004zero}, and FRI \cite{goswami1999postural} are classified into postural stability methods, among which ZMP is more common than others.

The nominal gaits achieved from postural/cyclic stability methods are most\-ly marginally stable such that the slightest disturbance may often cause the robot to fall.
In  the confrontation with slight disturbances, the joint trajectory tracking control of the nominal gait may preserve stability, but against significant ones, the tracking gait has to be adjusted to re-establish the conditions of stability.
For ZMP-based gaits, the gait modification is performed through ZMP compensation. Methods such as injecting a compensation torque into the ankle joint \cite{prahlad2008disturbance}, using the rotary momentum of non-contact/upper limbs \cite{hill2015active}, and support polygon adjustment via foot placement \cite{yu2018disturbance} are among proposed solutions in the literature for this purpose.  
On the other hand, the gait evolution can be investigated in a discrete event-based scheme. A discrete map is established between specific events (\eg\ the beginning of each walking step), and gait variables get modified at those instances. These intermittent modifications are applied through joint space inputs \cite{miura1984dynamic} or by specifying the gait parameters (\eg, step length and period \cite{de2012foot,khadiv2020walking}) at each step, or even by opting for an open-loop strategy such as the swing-leg retraction method \cite{hobbelen2008swing}. Despite triggering the discrete controller at specified time instances, the method is extendable to provide immediate corrections at every intermediate time by projecting the measured error values back in time and obtaining equivalent discrete-time control inputs \cite{faraji2019bipedal}.

One of the hazardous circumstances that can completely affect the effectiveness of a stabilizer is the odds of slipping, commonly encountered in unknown environments such as outdoor conditions. 
Several studies have been conducted on generating gaits for bipeds walking on slippery surfaces, pertaining to two categories: preventing slippage before it occurs or recovering from it. The problem is mainly dealt with in the absence of significant disturbances.
Slippage is preventable beforehand by planning the gait in a way that the corresponding required friction coefficient is less than what the surface can provide. This condition can be imposed as a hard constraint to the optimization problem at a high level in the footstep planning algorithm \cite{brandao2016footstep}, or, more conventionally, at a low level in the joints trajectory \cite{feng20133d}. The risk of slippage can be reduced by defining a gait-planning optimization problem that minimizes the horizontal acceleration of the center of mass (CoM) as a part of its objective \cite{kajita2004biped}.
By extending gait planning optimization problems to account for stick-slip transitions, pre-planned gaits which incorporate slippage can be generated to provide stable walking on low-friction surfaces \cite{ma2019dynamic}.
Using reflex strategies such as lifting the hip for the immediate modification of ground reaction forces \cite{park2001reflex}  is another way used for slippage recovery  on low-friction surfaces. 
Besides, employing certain strategies, such as the swing-leg retraction, can reduce the minimum friction requirement while maintaining stability and may even help to tolerate some slipping without risk of falling \cite{chen2021robust}.

As previously alluded, although few studies investigated biped walking on slippery surfaces, nonetheless none took into account the friction limitation of the surface to adapt the gait while simultaneously confronting external disturbances. 
By following this trend, it is expected to achieve more reliability in biped walking, as well as higher functionalities such as push recovery and gait switching ability on low-friction surfaces.
In this paper, a discrete event-based gait longitudinal motion controller is developed that adjusts both gait parameters (step length and time period) at the beginning of each step according to changing circumstances and surface friction limitations. For this purpose, the dynamical model of the biped robot is extracted in the task space, in terms of the CoM variables. Then, a no-slippage safe region is determined in the state space such that keeping the footstep initial state in that region prevents the biped from slipping. By taking into account the safe region as a constraint of motion, a preliminary step length controller emerged as the motion stabilizer. In confronting situations where the initial state is out of the safe region, a step-time adjustment scheme is considered for preventing  slippage. The integration of both step-length and step-time adaptation methods leads to a robust motion stabilizer. The soundness of the proposed algorithm is formally proven in the sagittal plane and also validated through several numerical simulations performed under different surface conditions, with scenarios involving push recovery and switching between different walking gaits.

The rest of the paper is organized as follows. In Section~2, the task space mathematical model of biped is extracted. Section~3 presents the no-slipping safe region. The step-length and step-time controllers are developed in Sections~4 and 5, respectively. Section~6 reports simulation results obtained by their combination. A discussion and comparative study on the advantages and limitations of the proposed stabilizer comes in Section~7, followed by the conclusion established in Section~8. The appendix provides  detailed proofs of all theoretical propositions.


\section{Background}
\label{sec:back}

\subsection{Biped Walking}
\label{sec:bipedWalking}

\begin{figure}
	\centering
	\includegraphics{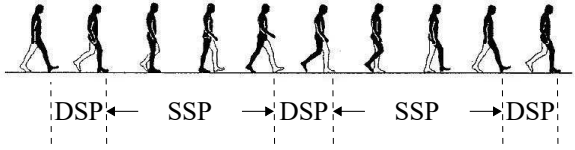}
	\caption{A schematic of biped walking containing two main phases: SSP and DSP. Adapted from \cite{inman1981human}.}
	\label{fig:WalkingPhases}
\end{figure}

Biped walking is defined as "moving along at a moderate pace by lifting up and putting down each foot in
turn, so that one foot is on the ground while the other is being lifted" \cite{kajita2014introduction}.
Therefore, one foot is at least in contact with the ground during walking.
Accordingly, walking consists of two main phases: a double support phase (DSP) and a single support phase (SSP).
In DSP, both feet are in contact with the ground, whereas in SSP, one foot is in contact while the other one is lifted (see Fig.~\ref{fig:WalkingPhases}).
The leg in touch is called the stance/support leg, and the other one is called the swing leg. 

\subsection{Mathematical Model}

In this paper, we consider the dynamics of the biped robot equivalently projected at its center of mass (CoM), called centroidal dynamics \cite{orin2013centroidal}.
The centroidal dynamics describes the interaction of the robot and the environment without dealing with the  dynamics in the joint space.
Thus, biped walking kinetics can be expressed with respect to some generalized coordinates (\eg,  the overall linear and angular momentum of the robot) and the reaction forces of the ground.
In this section, the centroidal dynamics of a biped in SSP and DSP modes is obtained.

\begin{figure}
	\centering
	\subfloat[SSP]{
		\includegraphics{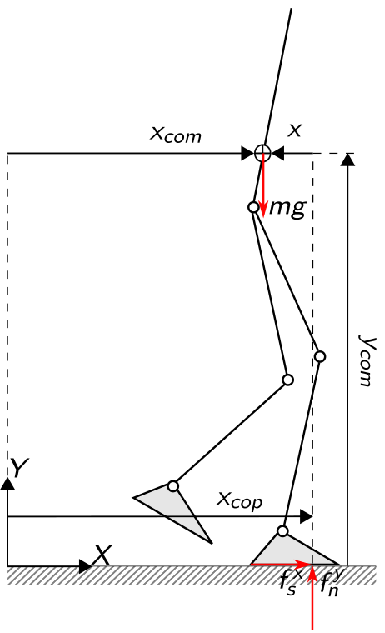}
		\label{fig:biped_SSP}
	}
	\subfloat[DSP]{
		\includegraphics{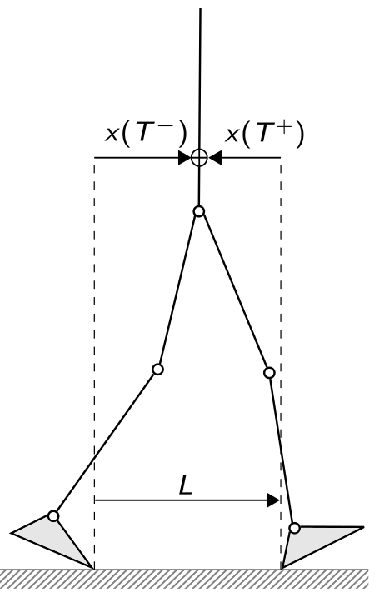}
		\label{fig:biped_DSP}
	}
	\caption{The diagram of a conventional biped in \protect\subref{fig:biped_SSP} SSP and \protect\subref{fig:biped_DSP} DSP configuration}
	\label{fig:biped}
\end{figure}

\subsubsection{Centroidal Dynamics in SSP}

A biped in SSP is depicted in Fig.~\ref{fig:biped_SSP}.
Accordingly, the governing dynamic equations in this phase are computed using Newton's second law as follows
\begin{align}
&f_s^x=m\ddot{x}_{com}\label{eq:firstmotioneq},\\
&f_n^y-mg=m\ddot{y}_{com}\label{eq:secondmotioneq},\\
&f_s^x y_{com} - f_n^y (x_{com}-x_{cop}) = \dot{H}\label{eq:thirdmotioneq},
\end{align}
where $m \in \mathbb{R}^+$ is the total mass of the biped, $g \in \mathbb{R}^+$ is the gravity acceleration, $x_{com}, y_{com} \in \mathbb{R}$ are the horizontal and vertical position of CoM, respectively. $x_{cop} \in \mathbb{R}$ is the center of pressure (CoP) of the reaction forces applied by the ground to the robot.
$H\in\mathbb{R}$ is the overall angular momentum of the biped around its CoM, and $f_s^x, f_n^y \in \mathbb{R}$ are the horizontal and vertical components of the reaction forces, respectively.

According that the CoM vertical motion may be considered negligible in comparison to its horizontal component, its overall motion can be restricted to the horizontal direction. The following assumptions are also considered:
\begin{assumption}
	For the biped shown in Fig.~\ref{fig:biped}, we assume that
	\begin{enumerate}[{A1}-1]
		\item The acceleration of the CoP is zero, \ie, $\ddot{x}_{cop} = 0$.
		\item The  height of the CoM  is constant, \ie, $\ddot{y}_{com} = 0$.
		\item $H$ is constant, \ie, $\dot{H}=0$.
	\end{enumerate}
\end{assumption}

Considering the above assumptions, Eqns.~\eqref{eq:firstmotioneq}-\eqref{eq:thirdmotioneq} can be simplified as
\begin{equation}
\label{eq:SimplifiedModel}
	\ddot{x}=\omega^2 x,
\end{equation}
where $x = x_{com} - x_{cop}$, $\omega=\sqrt{\frac{g}{h}}$, and $h \in \mathbb{R}^+$ denotes the CoM height.

\subsubsection{Centroidal Dynamics in DSP}

At the moment the swing foot touches the ground, the single support phase ends and the double support phase begins, as depicted in Fig.~\ref{fig:biped_DSP}.
The following assumptions are considered for modeling DSP. 

\begin{assumption}
	For the biped, we assume that
	\begin{enumerate}[{A2}-1]
		\item The DSP is instantaneous, \ie, the stance foot lifts upon the swing foot touches the ground.
		\item The swing foot lands at zero velocity on the ground and so there is no impact. \label{asm:zeroVelocity}
		\item The stance foot lifts from the ground without scuffing.
		\item The CoP is shifted instantaneously by a step length at each phase transition, \ie, the CoP is always under one of the soles, not somewhere between.
	\end{enumerate}
\end{assumption}

Regarding the above assumption, we have
\begin{equation}
\label{eq:DSPdynamics}
\begin{aligned}
	&x(T^+)=x(T^-)-L,\\
	&\dot{x}(T^+)=\dot{x}(T^-),
\end{aligned}
\end{equation}
where superscripts $\cdot^-$ and $\cdot^+$ denote the instants before and after the DSP.
$T \in \mathbb{R}^+$ is the moment when the DSP occurs, and $L \in \mathbb{R}$ is the step length.

\subsubsection{Overall Discrete Dynamics}

The SSP dynamics~\eqref{eq:SimplifiedModel} is a linear second-order differential equation.
Thus, considering $\left[ x_0^i, \dot{x}_0^i\right]$ as the initial condition of the step $i$ and $T(i)$ as its time duration, the solution of~\eqref{eq:SimplifiedModel} for the step $i$ is obtained as
\begin{equation}\label{eq:XSolution}
x^i(t) = \left( \frac{e^{\omega t} + e^{ - \omega t}}{2} \right) x_0^i+ \left( \frac{e^{\omega t} - e^{ - \omega t}}{2\omega } \right)\dot x_0^i,
\end{equation}
where $x^i$ denotes $x$ during the step $i$, and $t\in[0,T(i)]$.

On the other hand, one can rewrite \eqref{eq:DSPdynamics} for the DSP between the consecutive steps $i$ and $i+1$ as follows
\begin{equation}
\begin{aligned}
	&x_0^{i+1} = x^i(T(i))-L(i),\\
	&\dot{x}_0^{i+1} = \dot{x}^i(T(i)),
\end{aligned}
\end{equation}
where $L(i)$ is the step length of  $i$th step.
Substituting \eqref{eq:XSolution} in the above equations, the relation between the initial conditions of two consecutive steps is computed as

\begin{equation}
\label{eq:MapEquation}
	\mathbf{x}_0^{i+1} =\mathbf{A}(T(i))\mathbf{x}_0^i + \mathbf{b}L(i),
\end{equation}
where $\mathbf{x}_0^i = \begin{bmatrix} x_0^i, \dot{x}_0^i \end{bmatrix}^\top$ is the initial condition for the $i$th step, and
\begin{gather}
\mathbf{A}(T(i))=\frac{1}{2}
\begin{bmatrix}
e^{\omega T(i)}+e^{-\omega T(i)} & \frac{e^{\omega T(i)}-e^{-\omega T(i)}}{\omega}\\[10pt]
\omega \left( e^{\omega T(i)}-e^{-\omega T(i)} \right) & e^{\omega T(i)}+e^{-\omega T(i)}
\end{bmatrix}
,\\[5pt]
\mathbf{b}=\begin{bmatrix}
-1\\
0
\end{bmatrix}.
\end{gather}

\begin{property}\label{prop:AMatrixProperties}
For the matrix $\mathbf A$, one can show that
\begin{enumerate}[{P1}-1]
    \item $A_{11}, A_{12}, A_{21}, A_{22} \in \mathbb{R}^+$,
	\item $A_{11}=A_{22}$, and
	\item $A_{11}^2-A_{12}A_{21}=1$.
\end{enumerate}
\end{property}

The difference equation \eqref{eq:MapEquation} is a discrete-time dynamical system (Poincar\'e map) where $\mathbf{x}_0$ is the state vector, and $L$ and $T$ constitute its inputs.
The fixed point of equation  \eqref{eq:MapEquation} indicates the initial condition that leads to a periodic solution for the overall system \eqref{eq:SimplifiedModel} and \eqref{eq:DSPdynamics}.
For a fixed step-length and step-time denoted by $L^*$ and $T^*$, respectively, the fixed point of \eqref{eq:MapEquation}, $\mathbf{x}_0^*$, becomes 
\begin{equation}
\mathbf{x}^*_0 = \left(I_2 - \mathbf{A}(T^*) \right)^{-1} \mathbf{b}L^* =
\frac{L^*}{2}
\begin{bmatrix}
-1\\
\frac{{{e^{\omega {T^*}}} + 1}}{{{e^{\omega {T^*}}} - 1}}\omega 
\end{bmatrix}.
\end{equation}
Accordingly, for $\left[ x(0),\dot{x}(0) \right]^\top = \mathbf{x}_0^*$, the solution of the overall system \eqref{eq:SimplifiedModel} and \eqref{eq:DSPdynamics} is a periodic walking with step length $L = L^*$ and step time $T = T^*$.
For the purpose of stability analysis of this special periodic motion at constant step time $T = T^*$ and step length $L = L^*$, the behaviour of its discrete dynamics is studied under a perturbed initial condition, \ie, $\mathbf{x}_0^i=\mathbf{x}_0^*+\Delta \mathbf{x}_0^i $.
According to \eqref{eq:MapEquation}, the perturbed dynamics equation becomes
\begin{equation}\label{eq:PerturbationDynamics}
\Delta \mathbf{x}_0^{i+1} = \mathbf A(T^*) \Delta \mathbf{x}_0^i.
\end{equation}
The eigenvalues of $\mathbf A(T^*)$ are calculated as $e^{-\omega T^*}$ and $e^{\omega T^*}$.
Since $|e^{\omega T^*}|>1$, the perturbation dynamics \eqref{eq:PerturbationDynamics} is unstable about the fixed point $\Delta \mathbf{x}_0=\mathbf{0}$ \cite{strogatz2018nonlinear}. 
Thus, the fixed point $\mathbf{x}_0^*$ is unstable at fixed step length and time.
Considering \eqref{eq:MapEquation}, one can alleviate these conditions by employing a variable step length $L(i)$ and/or step time $T(i)$ for stabilizing the fixed point $\mathbf{x}_0^*$.
In this paper, we initially present a step-length control technique for stabilizing the fixed point $\mathbf{x}_0^*$ while ensuring that the stance foot does not slip during the walk.
For this purpose, the following assumption is considered.

\begin{assumption}\label{th:FixedTimeAssumption}
The step time of the whole walking process is set to a fixed desired value, \ie, $T(i)=T^*$ for all footsteps.
\end{assumption}
Considering the above assumption, the deviation $\Delta \mathbf{x}_0$ from the periodic solution, known as the error dynamics, is governed by 
\begin{equation}
\label{eq:ErrorDynamics}
\Delta \mathbf{x}_0^{i+1} = \mathbf{A}(T^*) \Delta \mathbf{x}_0^i + \mathbf{b} \Delta L (i),
\end{equation}
where $\Delta L(i)=L(i)-L^*$.
In the following, the aforementioned error dynamics will be used to design our intended step-length controller.


\section{Friction Analysis}
\label{sec:frictionAnalysis}

In this section, the non-slipping conditions for a biped in SSP is investigated using the concept of static friction.
Then, a safe region is introduced as the set of initial states for the $i$th step (\ie, $\mathbf{x}_0^i = \left[x_0^i, \dot{x}_0^i \right]^\top$) that ensures non-slipping conditions to be fulfilled during the step.
Finally, a safe interval for the step length of the $i$th step (\ie, $L(i))$ is obtained by guarantying $\mathbf{x}_0^{i+1}$ remains in the safe region provided that $\mathbf{x}_0^{i}$ starts there.

\subsection{Non-Slipping Condition}

Let's consider the biped pushing off the ground with its stance foot for providing a proper reaction that drives it forward.
Based on the Coulomb friction model  \cite{erdmann1994representation}, the stance foot does not slip if $|f_s^x| < \mu f_n^y$, where $\mu \in \mathbb{R}^+$ is the coefficient of static friction.
The stance foot is prone to slip if $|f_s^x| =\mu f_n^y$.
Consequently, the stance foot will not slip if
\begin{equation}
    \label{eq:nonSlipCondition}
    \max_t\left|f_s^x (t)\right|<\mu f_n^y.
\end{equation}
Above inequality is termed as the non-slipping condition.

According to \eqref{eq:firstmotioneq} and \eqref{eq:SimplifiedModel}, the left-hand side of the non-slipping condition can be rewritten as
\begin{equation}
\label{eq:leftnsc}
\max_t \left|f_s^x(t)\right| = m \omega^2 \max_t|x(t)|.
\end{equation}
According to \eqref{eq:secondmotioneq}, the right-hand side of the non-slipping condition can be rewritten as
\begin{equation}
\label{eq:rightnsc}
\mu f_n^y = \mu m g.
\end{equation}
Considering \eqref{eq:leftnsc} and \eqref{eq:rightnsc}, the non-slipping condition \eqref{eq:nonSlipCondition} is simplified as 
\begin{equation}
\label{eq:nsdistance}
\max_t |x(t)| < \dfrac{\mu g}{\omega ^2} = \mu h.
\end{equation}
As a result, the non-slipping condition \eqref{eq:nonSlipCondition} expressed at the force level is now related to kinematic terms \eqref{eq:nsdistance}.
Besides the reduced complexity, following interpretations can be deduced:
\begin{enumerate}
    \item If the horizontal distance between the CoM of the body and the CoP remains less than $\mu h$, the stance foot will not slip.
    \item Both $\mu$ and $h$ have a direct effect on reducing the risk of slipping, one by providing a sufficient shear resistance against slipping, the other by reducing the dynamic force induced by the swinging leg.
\end{enumerate}

\subsection{Safety Region Determination for Initial Conditions}
\label{sec:safeRegion}

\begin{figure}
 \centering
 \includegraphics{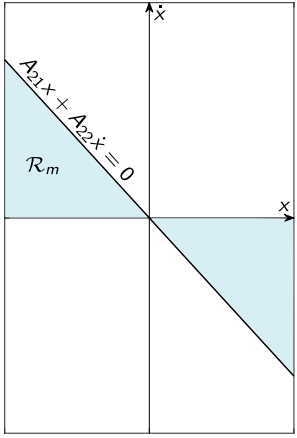}
 \caption{A schematic of region $\mathcal{R}_m$} 
 \label{fig:Rm}
\end{figure}

According that the simplified model of the walking process, \eqref{eq:SimplifiedModel}, is represented as an autonomous system, the time evolution of $x$ only depends on its initial state.
Therefore, one can express the non-slipping condition \eqref{eq:nsdistance} of a step solely in terms of its initial state.
In this way, the set of initial states that ensures the non-slipping condition during the step (defined as the safe region)
is computed as follows
\begin{equation}
\label{eq:safeRegionDef}
    \mathcal{S} = \left\{ [x_0,\dot{x}_0] \, : \, \max_{0 \leq t \leq T} |x(t)| < \mu h \right\},
\end{equation}
where $[x_0,\dot{x}_0] = [x(0),\dot{x}(0)]$ is the initial state of the step, and $T$ is its period.
As $x(t)$ is continuous, the maximum value of $x(t)$ in a whole step can occur at an interior point or at the extremities:
\begin{enumerate}
    \item $x_0 = x(0)$,
    \item $x_T = x(T)$, or
    \item $x_m=x(t_m)$ as the extremum of $x(t)$ occurring in the interval.
\end{enumerate}
where the instant $t_m$ is obtained through derivating \eqref{eq:XSolution} with respect to time, which leads to
\begin{equation}
\label{eq:tm}
t_m=\frac{1}{2\omega}\ln{\frac{\omega x_0-\dot{x}_0}{\omega x_0+\dot{x}_0}},
\end{equation}
and thus
\begin{equation}\label{eq:xm}
x_m = \pm\omega^{-1}\sqrt{(\omega x_0-\dot{x}_0)(\omega x_0+\dot{x}_0)}.
\end{equation}
Note that $x_m$ is acceptable iff $x_m$ occurs within the current step interval, \ie, $t_m \in (0,T)$.
Considering \eqref{eq:tm}, $t_m \in (0,T)$ iff
\begin{equation}
\dot x_0(\dot x_0+\frac{A_{21}}{A_{22}}x_0)<0.
\end{equation}
The set of initial states $[x_0,\dot{x}_0]$ satisfying the above inequality is denoted by $\mathcal{R}_m$.
Thus, $x_m$ is the extremum of $x$ iff $[x_0,\dot{x}_0] \in \mathcal{R}_m$.
Figure~\ref{fig:Rm} shows $\mathcal{R}_m$ in the state space $(x,\dot{x})$.

\begin{figure*}
 \centering
 \subfloat[]{\includegraphics{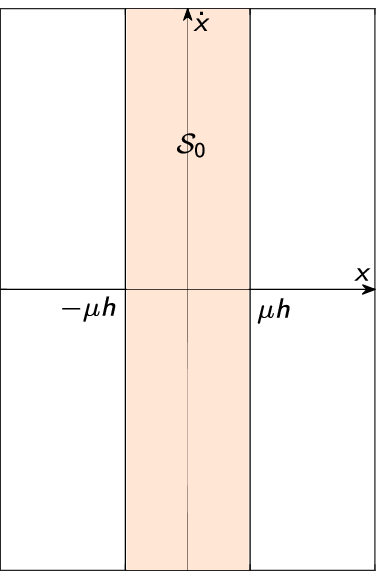}\label{fig:S0}}~
 \subfloat[]{\includegraphics{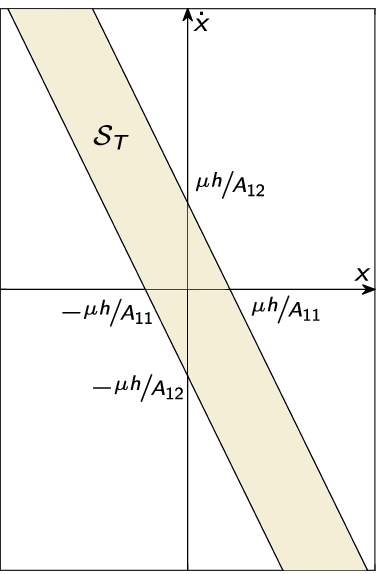}\label{fig:ST}}~
 \subfloat[]{\includegraphics{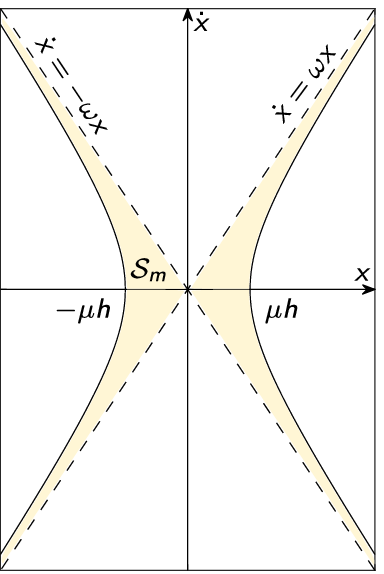}\label{fig:Sm}}\\
 \subfloat[]{\includegraphics{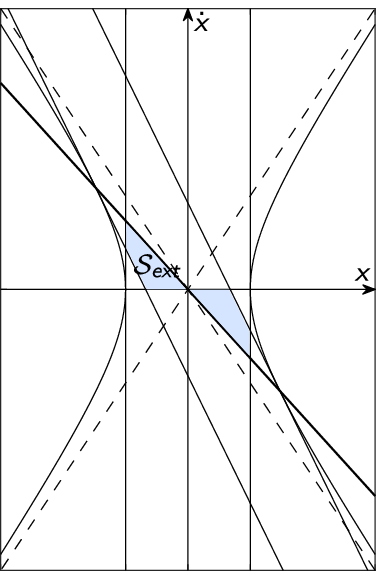}\label{fig:Sext}}~
 \subfloat[]{\includegraphics{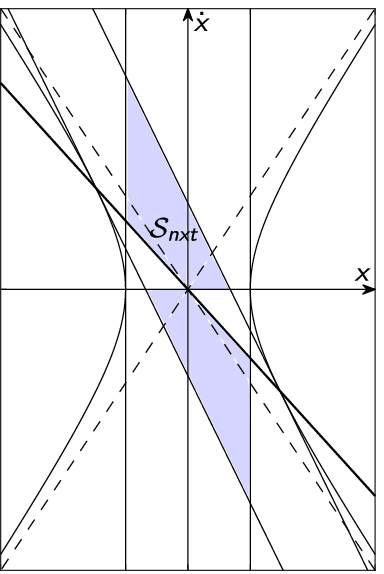}\label{fig:Snxt}}~
 \subfloat[]{\includegraphics{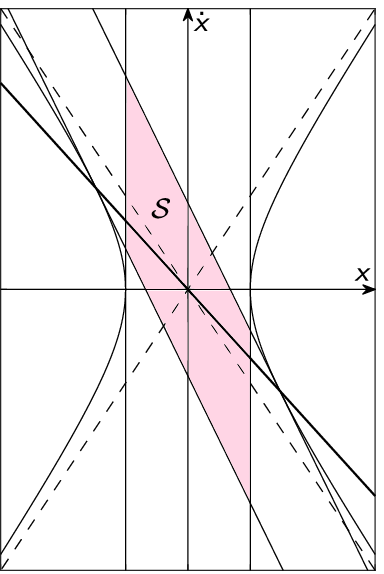}\label{fig:S}}
 \caption{Schematics of safe regions: 
 \protect\subref{fig:S0} $\mathcal{S}_0$;
 \protect\subref{fig:ST} $\mathcal{S}_T$;
 \protect\subref{fig:Sm} $\mathcal{S}_m$;
 \protect\subref{fig:Sext} $\mathcal{S}_{ext}= \mathcal{S}_0\cap \mathcal{S}_T\cap \mathcal{S}_m\cap \mathcal{R}_m$;
 \protect\subref{fig:Snxt} $\mathcal{S}_{nxt}=\mathcal{S}_0 \cap \mathcal{S}_T\cap \mathcal{R}_m^c$;
 \protect\subref{fig:S} $\mathcal{S}= \mathcal{S}_{ext} \cup \mathcal{S}_{nxt}$. 
 The border of $\mathcal{R}_m$ is shown thicker in the bottom plots.}
 \label{fig:safeRegion}
 \end{figure*}

According to the above explanations, the non-slipping condition \eqref{eq:safeRegionDef} can be expressed as
\begin{equation}\label{eq:ineqs}
\mathcal{S} = \begin{dcases}
\max \left\{|x_0|,|x_m|,|x_T|\right\} < \mu h, & \mathbf x_0\in \mathcal{R}_m \\
\max\left\{|x_0|,|x_T|\right\} < \mu h, & \mathbf x_0\notin \mathcal{R}_m
\end{dcases}\\
\end{equation}
Accordingly, three regions $\mathcal{S}_0$, $\mathcal{S}_m$ and $\mathcal{S}_T$ can be defined as
\begin{align}
&\mathcal{S}_0 = \left\{ [x_0,\dot{x}_0] \, : \, |x_0| < \mu h \right\}, \label{eq:S0}\\
&\mathcal{S}_m = \left\{ [x_0,\dot{x}_0] \, : \, |x_m| < \mu h \right\},\\
&\mathcal{S}_T = \left\{ [x_0,\dot{x}_0] \, : \, |x_T| < \mu h \right\},
\end{align}
so that $\mathcal{S}_0$, $\mathcal{S}_m$ and $\mathcal{S}_T$ are the sets of initial states ensuring no slippage occurrence, respectively at the start, during and at the end of the footstep.
Considering \eqref{eq:XSolution}, the following relation is obtained, $x_T = A_{11}x_0 + A_{12} \dot{x}_0$, and thus
\begin{equation}\label{eq:STComputation}
    \mathcal{S}_T = \left\{ [x_0,\dot{x}_0] \, : \, \frac{-A_{11}x_0-\mu h }{A_{12}}< \dot{x}_0 < \frac{-A_{11}x_0+\mu h}{A_{12}} \right\}.
\end{equation}
According to \eqref{eq:xm}, $\mathcal{S}_m$ can be rewritten as
\begin{equation}
    \mathcal{S}_m = \left\{ [x_0,\dot{x}_0] \, : \, 0< \omega^2x_0^2-\dot{x}_0^2< (\omega \mu h )^2 \right\}.
\end{equation}
The regions $\mathcal{S}_0$, $\mathcal{S}_m$ and $\mathcal{S}_T$ are shown in Figs.~\ref{fig:S0} to \ref{fig:Sm}.
Therefore, the non-slipping condition \eqref{eq:ineqs} can be rewritten as
\begin{equation}
\mathcal{S} = \begin{dcases}
\label{eq:SextSnxt}
\mathcal{S}_{ext} = \mathcal{S}_0\cap \mathcal{S}_T\cap \mathcal{S}_m\cap \mathcal{R}_m,\\
\mathcal{S}_{nxt} = \mathcal{S}_0 \cap \mathcal{S}_T\cap \mathcal{R}_m^c,
\end{dcases}
\end{equation}
where $\mathcal{R}_m^c$ is the complementary of $\mathcal{R}_m$. $S_{ext}$ and $S_{nxt}$ are shown in Figs.~\ref{fig:Sext} and \ref{fig:Snxt}. Their ensemble constitute the entire safety region, $\mathcal{S}$, shown in Fig.~\ref{fig:S}, and represented as
\begin{equation}
\label{eq:safeRegion}
    \mathcal{S} = \mathcal{S}_{ext} \cup \mathcal{S}_{nxt}.
\end{equation}

In conclusion, the  condition for taking a non-slipping step is related to respect a corresponding safe region for the initial state of the body's CoM.
In this manner, the stance foot will not slip during the $i$th step if $[x_0^i,\dot{x}_0^i] \in \mathcal{S}$.

\begin{remark}\label{rem:SStructure}
Figure.~\ref{fig:safeRegion} results that $\mathcal{S}$ is at the intersection of two regions $\mathcal{S}_0$ and $\mathcal{S}_T$, \ie, $\mathcal{S}=\mathcal{S}_0\cap{S}_T$.
\end{remark}

\subsection{Safe Range for Step Length}

As discussed in section \ref{sec:safeRegion}, the stance foot of the robot does not slip during the $i$th step if $[x_0^i,\dot{x}_0^i] \in \mathcal{S}$.
Assuming $[x_0^i,\dot{x}_0^i] \in \mathcal{S}$, one can compute the step length $L(i)$ ensuring that $[x_0^{i+1},\dot{x}_0^{i+1}] \in \mathcal{S}$.

According to Remark~\ref{rem:SStructure}, $\left[ x_0^{i+1},\dot{x}_0^{i+1} \right] \in \mathcal{S}$ iff the following conditions hold
\begin{enumerate}
	\item $\left[ x_0^{i+1},\dot{x}_0^{i+1} \right] \in \mathcal{S}_0$, and
	\item $\left[ x_0^{i+1},\dot{x}_0^{i+1} \right] \in \mathcal{S}_T$.
\end{enumerate}
Considering the discrete dynamics \eqref{eq:MapEquation}, the step length $L(i)$ ensuring the above conditions shall satisfy the following
\begin{enumerate}
\item 
$\left[ x_0^{i+1},\dot{x}_0^{i+1} \right] \in \mathcal{S}_0$ iff $L(i)\in\left(L_l^0(i),L_u^0(i)\right)$ where
\begin{gather}
L_l^0(i) = A_{11}x_0^i+A_{12}\dot{x}_0^i-\mu h,\label{eq:Ll0}\\
L_u^0(i) = A_{11}x_0^i+A_{12}\dot{x}_0^i+\mu h.\label{eq:Lu0}
\end{gather}
\item
$\left[ x_0^{i+1},\dot{x}_0^{i+1} \right] \in \mathcal{S}_T$ iff $L(i)\in\left(L_l^T(i),L_u^T(i)\right)$ where
\begin{gather}
L_l^T(i) = \left(2A_{11}-\frac{1}{A_{11}}\right)x_0^i+2A_{12}\dot{x}_0^i-\frac{\mu h}{A_{11}},\label{eq:LlT}\\
L_u^T(i)=\left(2A_{11}-\frac{1}{A_{11}}\right)x_0^i+2A_{12}\dot{x}_0^i+\frac{\mu h}{A_{11}}.\label{eq:LuT}
\end{gather}
\end{enumerate}
According to the above explanations, $\left[ x_0^{i+1},\dot{x}_0^{i+1} \right]\in\mathcal{S}$ iff ${L(i) \in \left( L_l^s(i),L_u^s(i) \right)}$ where
\begin{gather}
L_l^s(i)= \max \left\{ L_l^0(i),L_l^T(i) \right\},\label{eq:LlSafe}\\
L_u^s(i)= \min \left\{ L_u^0(i),L_u^T(i) \right\}.\label{eq:LuSafe}
\end{gather}

In conclusion, the stance foot of the robot does not slip during the $i$th footstep if
\begin{enumerate}
	\item $\mathbf{x}_0^i\in\mathcal{S}$, and
	\item $L(i) \in \left(L_l^s(i),L_u^s(i)\right)$.
\end{enumerate}
We call $\left(L_l^s(i),L_u^s(i)\right)$ the step-length safe range.

\begin{proposition}\label{th:SafeRange}
	If $\mathbf{x}_0^i \in \mathcal{S}$ and $L(i) \in \left( L_l^{s}(i),L_u^s(i) \right)$, the safe range for the next step is nonempty, \ie, $\left( L_l^{s}(i+1),L_u^s(i+1) \right) \neq \varnothing$.
\end{proposition}
The proof is given in \ref{prf:Existance}.

According to the above proposition, regardless of the value for the step length selected within the safe range, a next safe interval always exists.


\section{Step Length Controller}

In this section, a step length control algorithm is designed for walking stabilization of the biped with given step length and period, $L^*$ and $T^*$.
Specifically, the proposed control algorithm  on step length is such that
\begin{enumerate}
	\item The initial states series converges to the desired value  $\mathbf{x}_0^*$, and
	 \item The state always remains within the safety region, i.e., $\forall i, \, \mathbf{x}_0^i \in \mathcal{S}$.
\end{enumerate}
In the first part of this section, considering the dynamics \eqref{eq:ErrorDynamics}, a control rule $\Delta L(i)$ is designed for stabilizing the cyclic solution around $\Delta \mathbf{x}_0 = \mathbf{0}$.
Then, a technique is presented for ensuring that $\forall i, \, \mathbf{x}_0^i \in \mathcal{S}$.

\begin{proposition}\label{th:ConvergenceRange}
	For the dynamical system \eqref{eq:ErrorDynamics}, if $\forall i$,
	\begin{equation}\label{eq:convergenceRange}
	\Delta L(i) \in \big(\min \{\Delta L_1(i), \Delta L_2(i)\},\max\{\Delta L_1(i), \Delta L_2(i)\}\big)
	\end{equation}
	where
	\begin{gather}
	\Delta L_1(i)=\frac{\splitfrac{\left(A_{11}A_{21}+A_{21}A_{22}+A_{21}\right)\Delta x_0^i}{+\left(A_{12}A_{21}+A_{22}A_{22}+A_{22}\right)\Delta \dot{x}_0^i}}{A_{21}},\label{eq:ConvRoot1} \\
	\Delta L_2(i)=\frac{\splitfrac{\left(A_{11}A_{21}+A_{21}A_{22}-A_{21}\right)\Delta x_0^i}{+\left(A_{12}A_{21}+A_{22}A_{22}-A_{22}\right)\Delta \dot{x}_0^i}}{A_{21}},\label{eq:ConvRoot2}
	\end{gather}
	then $\Delta \mathbf{x}_0 = \mathbf{0}$ is globally asymptotically stable.
\end{proposition}
The proof is given in \ref{prf:Convergence}.

In other terms, the above proposition claims that the initial state converges to the desired value as long as $L(i)\in\left(L_l^c(i),L_u^c(i)\right)$ where
\begin{gather}\label{eq:ConvRange}
L_l^{c}(i)=\min \left\{\Delta L_1(i),\Delta L_2(i)\right\} + L^*,\\
L_u^{c}(i)=\max \left\{\Delta L_1(i),\Delta L_2(i)\right\} + L^*.
\end{gather}
$\left(L_l^c(i),L_u^c(i)\right)$ is called here the step-length range of convergence.

By considering an interval for the step length ensuring both safety and convergence, the safe-convergence range is defined as $\left(L_l^{sc}(i),L_u^{sc}(i)\right)$ where
\begin{gather}
L_l^{sc}(i)=\max \left\{L_l^{s}(i),L_l^{c}(i)\right\},\\
L_u^{sc}(i)=\min \left\{L_u^{s}(i),L_u^{c}(i)\right\}.
\end{gather}
Accordingly, if $\forall i, \, L(i) \in \left(L_l^{sc}(i),L_u^{sc}(i)\right)$, then the initial state converges to the desired value while satisfying the non-slipping condition provided that $\mathbf{x}_0^1\in\mathcal{S}$.
The following proposition implies the existence of a safe-convergence range.

\begin{proposition}\label{th:SafeConvRangeIntersection}
The intersection of the step length safe range and convergence range is nonempty, \ie, $\left(L_l^s(i),L_u^s(i)\right)\cap\left(L_l^c(i),L_u^c(i)\right)\neq \varnothing$.
\end{proposition}
The proof is given in \ref{prf:Intersection}.

According to the above proposition and proposition~\ref{th:SafeRange}, if $\mathbf{x}_0^i\in\mathcal{S}$ and $L(i)$ is in the safe-convergence range of the step $i$, then the safe-convergence range for the step $i+1$ is nonempty.
For the sake of simplicity, let's select
\begin{equation}\label{eq:LengthController}
L(i)=\frac{1}{2}\left(L_l^{sc}(i)+L_u^{sc}(i)\right).
\end{equation}


\section{Step Time Controller}

The aforementioned step length controller can stabilize motion only if the condition $\mathbf{x}_0\in\mathcal{S}$ is fulfilled, otherwise slippage will occur which can not be prevented by the controller. In the case an initial state lying out of the safe region, two possibilities of slippage are conceivable as
\begin{enumerate}
\item
Immediate slippage when $\mathbf{x}_0\notin \mathcal{S}_0$, and
\item 
Subsequent slippage when $\mathbf{x}_0\notin \mathcal{S}_T$ but $\mathbf{x}_0\in \mathcal{S}_0$ \end{enumerate}
In the second case, slip occurs subsequently, for example after the intervention of an external pushing force, upon which the state leaves $\mathcal{S}_0$, \ie, the $x$ coordinate reaches the boundaries $\pm\mu h$.
In this scenario, slippage is preventable by adapting the step time accordingly. In this section, a step time controller is developed for preventing a subsequent slippage through two techniques of time adjustment, one with the particularity of establishing a safe region of "fixed-borders" and another one with "moving-borders".
Considering $\mathcal{S}_0$ and $\mathcal{S}_T$ definitions, \eqref{eq:S0} and \eqref{eq:STComputation}, a subsequent slippage occurs if $\mathbf{x}_0\in\mathcal{D}$ where
\begin{equation}\label{eq:DelayedSlipRegion}
\mathcal{D}=\{\mathbf{x}_0\mid|x_0|<\mu h,\text{ and }|A_{11}x_0+A_{12}\dot{x}_0|>\mu h\}.
\end{equation}
We call $\mathcal{D}$ the subsequent-slipping region. 

\subsection{Fixed-Border Time Adjustment}
\begin{figure}
 \centering
 \subfloat[]{\includegraphics{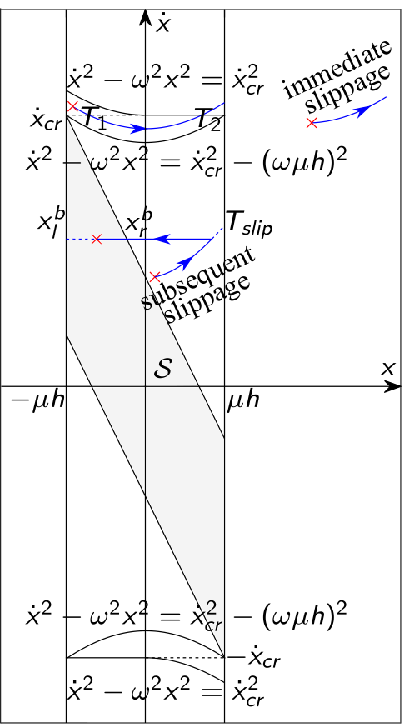}\label{fig:FixedBorder}}~
 \subfloat[]{\includegraphics{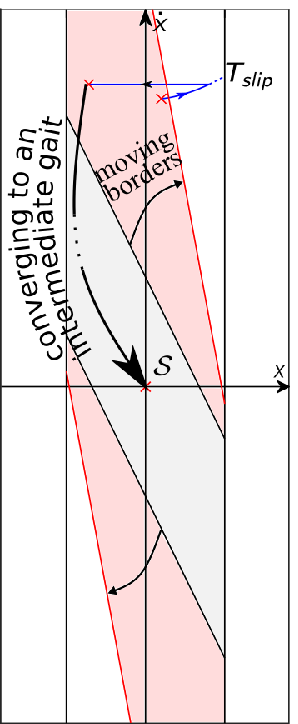}\label{fig:MovingBorder}}
 \caption{Time adjustment techniques: \protect\subref{fig:FixedBorder} fixed-border; \protect\subref{fig:MovingBorder} moving-border. The initial states are shown by (red) `$\times$' markers, and the safe regions $\mathcal{S}$ are colored.}
 \label{fig:FixedMovingBorder}
 \end{figure}

The basis of the method is illustrated  in Fig.~\ref{fig:FixedBorder}.
For a footstep starting from an initial state in $\mathcal{D}$, the main idea of the method is to prevent $x$ leaving $\mathcal{S}_0$ by reducing the step time and to return the next initial state into the safe region by a proper step length regulation. In this corrective step, converging to the desired gait is momentarily disregarded for assuring the necessary conditions of stability but is ultimately reestablished as a priority when back to safety. In fact, in this method, only the duration of the eminent-slipping step is adapted accordingly and will differ from the desired step time, thus preserving the safe region as before. In this regard, the method is nicknamed as "fixed-border" time adjustment. 
 
Considering \eqref{eq:XSolution}, the general form of equations in the state space $(x,\dot{x})$ is derived as
\begin{equation}
(\dot{x}^2-\dot{x}_0^2)-\omega^2(x^2-x_0^2)=0.
\end{equation}
Accordingly, slippage is preventable by the fixed-border time adjustment if $\mathbf{x}_0\in\mathcal{A}$ where
\begin{equation}\label{eq:FixedBorderA}
\mathcal{A}=\left\{\mathbf{x}_0\mid
\begin{aligned}
\big((\dot{x}_0^2-\omega^2 x_0^2<\dot{x}_{cr}^2)\cap( x_0\dot{x}_0<0)\big)\\
\cup\big((\dot{x}_0^2<\dot{x}_{cr}^2)\cap(x_0\dot{x}_0>0)\big)
\end{aligned}
\right\}\cap\mathcal{D},
\end{equation}
and $\dot{x}_{cr}=\frac{A_{11}+1}{A_{12}}\mu h$ (see Fig.~\ref{fig:FixedBorder}). In other words, if
\begin{enumerate}
\item
$\mathbf{x}_0\in\mathcal{A}$, then there exists a step time $T$ that prevents from subsequent slipping, and a step length $L$ to bring back the next initial state to $\mathcal{S}$.
\item
$\mathbf{x}_0\in\mathcal{D-A}$, then there exists a step time $T$ that prevents from subsequent slipping, but no step-length $L$ can bring back to safety $\mathcal{S}$.
\end{enumerate}

In the former circumstance, if $\mathbf{x}_0\in\mathcal{A}$, the state remains in $\mathcal{S}_0$ by selecting $T$ as
\begin{equation}
T\in
 \begin{dcases}
 (0,T_{slip}), &\text{if }\dot{x}_0^2-\omega^2 x_0^2\leq\dot{x}_{cr}^2-(\omega\mu h)^2\\
 (T_1,T_2), &\text{if }\dot{x}_0^2-\omega^2 x_0^2>\dot{x}_{cr}^2-(\omega\mu h)^2
 \end{dcases}
\end{equation}
where $T_{slip}\in\mathbb{R}^+$ is the slippage moment at which the $x$ trajectory intersects $\pm\mu h$, and $ T_1,T_2\in\mathbb{R}^+$ are the moments at which the $\dot{x}$ trajectory intersects $\pm\dot{x}_{cr}$ (see Fig.~\ref{fig:FixedBorder}); from \eqref{eq:XSolution}, one can calculate them as follows
\begin{gather}
T_{slip}=\frac{1}{\omega}\ln\dfrac{\omega\mu h+\sqrt{(\omega\mu h)^2+(\dot{x}_0^2-\omega^2 x_0^2)}}{|\dot x_0+\omega x_0|},\label{eq:TSlip}\\
T_1=\frac{1}{\omega}\ln\max\{1,\frac{\dot{x}_{cr}-\sqrt{\dot{x}_{cr}^2-(\dot{x}_0^2-\omega^2 x_0^2)}}{|\dot x_0+\omega x_0|}\},\\
T_2=\frac{1}{\omega}\ln\frac{\dot{x}_{cr}+\sqrt{\dot{x}_{cr}^2-(\dot{x}_0^2-\omega^2 x_0^2)}}{|\dot x_0+\omega x_0|}.
\end{gather}

Subsequently, the initial state of the next footstep is brought back into $S$ by selecting $L$ as
\begin{equation}\label{eq:FixedBorderStepLength}
L\in\left(x^i(T)-x^b_l,x^i(T)-x^b_r\right),
\end{equation}
where $x^b_l,x^b_r\in\mathbb{R}$ are the left and right extremity values of the safe range for $x$ (see Fig.~\ref{fig:FixedBorder}); Considering $\mathcal{S}$, both being calculated as follows
\begin{gather}
x^b_r=\min\left\{\mu h,\dfrac{\mu h-A_{12}\dot{x}^i(T)}{A_{11}}\right\},\\
x^b_l=\max\left\{-\mu h,\dfrac{-\mu h-A_{12}\dot{x}^i(T)}{A_{11}}\right\}.
\end{gather}

\subsection{Moving-Border Time Adjustment}
The basis of the method is illustrated in Fig.~\ref{fig:MovingBorder}. 
For an initial state $\mathbf{x}_0\in\mathcal{D}$, the main idea is to move the safe region borders by changing the desired step time to form a secondary safe region that includes $\mathbf{x}_0$ within itself.
Then, once the secondary safe region established, a new fixed-point (new cyclic gait) is selected as intermediate target between both primary and secondary safe regions. As the origin (a somehow marching-in-place gait) certainly belongs to both regions, it is selected as intermediate state.
Thereby, the stabilization process takes place in two stages;
First, the initial state is conducted from $\mathbf{x}_0$ to the intermediate state by the proper application of the step-length controller within the secondary safe region.
Next, the desired step time is restored to its original value, and convergence to the original fixed point (desired gait) is pursued by activating the step-length controller within the primary safe region. 
\begin{remark}
Upon the footstep initial state reaches the primary safe region during its convergence to the origin, the next stage can be proceeded right from there.
\end{remark}
The denomination of the method as moving-border time adjustment is due to the act of forming new borders to shape the safe region.
In order to determine the step time that will lead to the secondary safe region, let's consider the initial state $\mathbf{x}_0\in\mathcal{D}$ lying on the $\mathcal{S}_T$ border (26),  
\begin{equation}
A_{11}|_{@ T^*_m}x_0+A_{12}|_{@T^*_m}\dot{x}_0=\pm \mu h,
\end{equation}
 where $T^*_m\in\mathbb{R}^+$ is the new target step duration that results to the secondary safe region. The above equation provides the same solution as \eqref{eq:TSlip} for $T_m^*$. Therefore, to achieve the secondary safe region including $\mathbf{x}_0\in\mathcal{D}$, $T_m^*$ is opted as 
\begin{equation}
T^*_m\in(0,T_{slip}).
\end{equation}
Moreover, by setting the new desired step length equal to zero, the zero-gait (marching-in-place) state can be replaced as the intermediate fixed-point. 

\begin{remark}
Determining which time-adjustment techniques to select depends on the applicability of either one. Remind that the moving-border technique is applicable for the whole region $\mathcal{D}$, while the fixed-border technique applies only for $\mathcal{A}$ (which is a subset of $\mathcal{D}$). So any subset of $\mathcal{A}$ can be used as the triggering condition for the fixed-border approach, and accordingly its complementary for the other scheme.
\end{remark}


\section{Results}

In this section, the effectiveness of the developed controllers is demonstrated through performing some numerical simulations.

\subsection{Switching the Gait- Test against Different Ground Friction Values}\label{subsec:Switching}

\begin{figure*}
 \centering
 \subfloat[]{\includegraphics[height=9 in-43.7 pt]{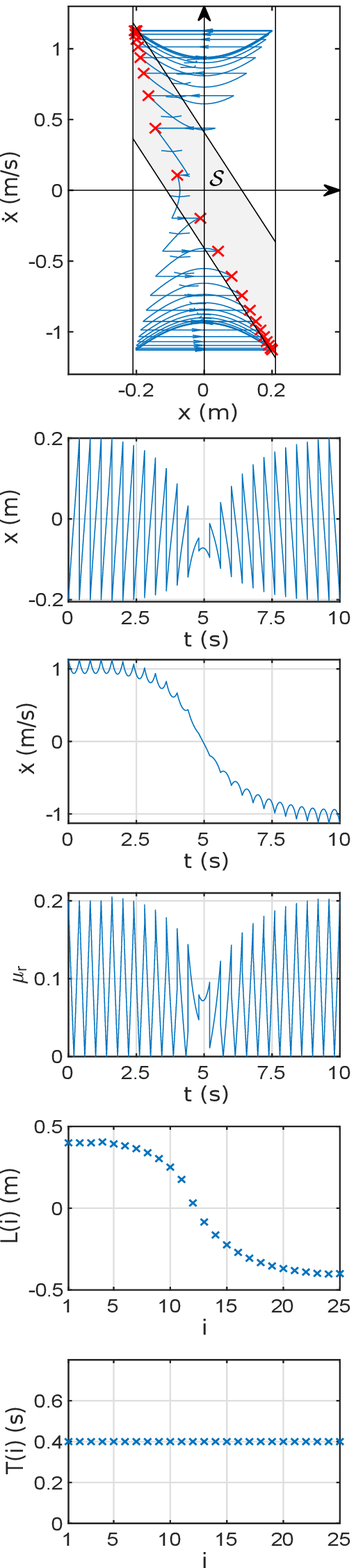}\label{fig:Switching21}}~
 \subfloat[]{\includegraphics[height=9 in-43.7 pt]{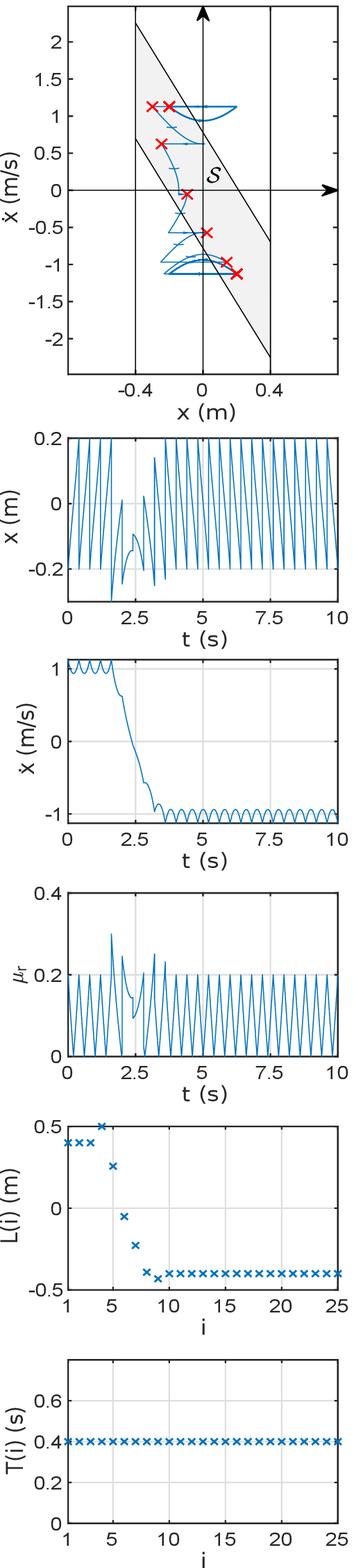}\label{fig:Switching4}}~
 \subfloat[]{\includegraphics[height=9 in-43.7 pt]{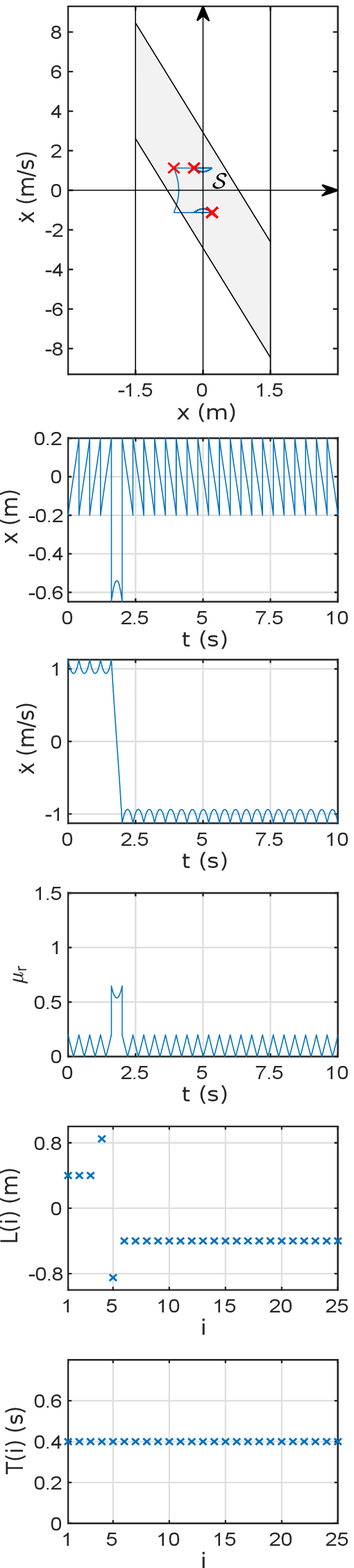}\label{fig:Switching15}}
 \caption{Switching between gaits on three surfaces with \protect\subref{fig:Switching21} $\mu=0.21$, \protect\subref{fig:Switching4} $\mu=0.4$, and \protect\subref{fig:Switching15} $\mu=1.5$. In the phase portraits (top subplots), limit cycles are shown thicker, the initial states by (red) `$\times$' markers, and the safe region $\mathcal{S}$ colored (in gray).}
 \label{fig:Switching}
\end{figure*}

Consider a biped postured in the sagittal plane with simplified dynamics  \eqref{eq:SimplifiedModel} and \eqref{eq:DSPdynamics}, walking from an initial state ${\mathbf{x}_0^*=[-0.2\text{ m},1.1274\text{ m/s}]^\top}$ with a step length $L^*=0.4\text{ m}$ and a step time $T^*=0.4\text{ s}$. The biped mass is $m=50 \text{ kg}$, and it tries to keep its CoM height at the constant value $h=1 \text{ m}$. At the onset of step $i=4$, the biped is ordered to change direction, i.e., to walk backward with the same step length and time. The results of simulating such a scenario on three different surfaces with $\mu=0.21$, $\mu=0.4$, and $\mu=1.5$, are depicted in Fig.~\ref{fig:Switching}.
 
As can be seen in the phase portraits, the walking trend switched from the forward gait (top limit cycle in the plot) to the backward gait (bottom limit cycle) while the footstep initial states always remained within the safe region $\mathcal{S}$. 
Accordingly, the friction coefficient required for ensuring the stability of this walking gait, denoted by $\mu_r$, remained lower than $\mu$ of the surface in all three cases (as $\mu_r$ plots show). 
However, as can be seen, the rate of convergence and the number of transient steps takes a different value for each case.

In the first case, Fig.~\ref{fig:Switching21}, where $\mu$ is just a little bigger than the required coefficient $\mu_r$ for the desired gait, alternating the motion is associated with some limitations. So the controller is forced to vary the step lengths in a restrictive manner (as $L(i)$ plot shows), resulting in slow convergence with more transient steps.

As $\mu$ increases, as in the second case, Fig.~\ref{fig:Switching4}, the controller is allowed to employ larger variations in step lengths, so the convergence rate increases as well, and the number of transient steps decreases.

In the third case, Fig.~\ref{fig:Switching15}, where the coefficient $\mu$ is largely sufficient in comparison to the require value $\mu_r$ for the desired gait, there are no worries of slippage, and thus the controller varies the step length as widely as necessary. So, the resulting convergence rate is maximum and the number of transient steps is minimal.

In fact, for walking on surfaces with friction coefficient $\mu$ less than the amount required for performing a particular gait, $\mu_r$, slippage is unavoidable. In the marginal case where $\mu$ equals $\mu_r$, reaching the desired gait itself without slipping is feasible but switching to another gait is not. In this case, the gait's initial state coincides with one of the boundaries $[\mp\mu h,\pm\dot{x}_{cr}]$ of the safe region.

Finally, as the initial states corresponding to the desired gaits lie within $\mathcal{S}$, switching between them only necessitates to modify the step length but step time adjustment is not required, as shown in $T(i)$ plots.

The simulation results show how the stabilizer adapts itself to different road situations. 
The results also reveal one of the advantages of this stabilizer over the state-of-art existing in the literature; 
Stabilization systems presented in \cite{de2012foot}, \cite{khadiv2020walking} and \cite{faraji2019bipedal} employ footstep adjustment techniques for confronting external disturbances, but are not expected to deal properly with the different aforementioned cases as none of them take into account the friction 
limitation of the surface to adapt the gait. 
Obviously, they will react uniformly to all cases, merely converging to the desired gait as fast as possible. This will induce exaggerated reactions for the first and second cases, leading to slippage and loss of stability.

From the results, we can infer that:
\begin{inparaenum}[(i)]
\item
The slipperiness of the surface depends on the maximum difference between the coefficients $\mu$ and $\mu_r$ related to the road and desired gait, respectively;
\item
The more slippery is the surface, the more restricted must be the variation of the walking parameters.  
\end{inparaenum}

\subsection{The Effect of Taller Height}

\begin{figure}
\centering
\includegraphics{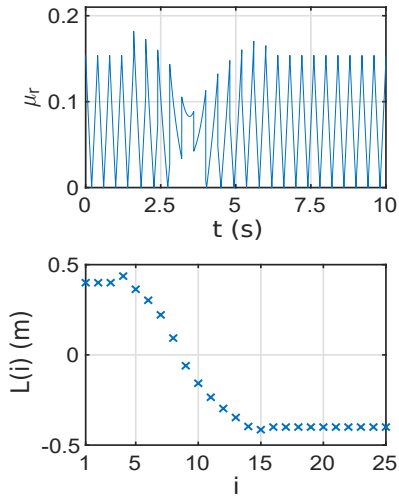}
\caption{The effect of a taller height on required friction coefficient and step length}
\label{fig:TallerHeight}
\end{figure}

Let's reconstitute the initial simulation performed in the previous section by employing the same parameters except for a taller height, varied to $h=1.3$ m. The simulation results are depicted in Fig.~\ref{fig:TallerHeight}.  
For brevity, only the required friction and step length variation plots are shown. As can be seen in the $\mu_r$ plot, the required friction has decreased compared to the similar plot depicted in Fig.~\ref{fig:Switching21}. Consequently, as the $L(i)$ plot indicates, the controller has been allowed to adapt to the situation by taking larger step lengths. This observation is expected from \eqref{eq:nsdistance} and confirms that increasing the CoM height can decrease the risk of slippage. In other words, \emph{taller persons are less prone to slippage in comparison to shorter ones when executing similar gaits}.
As mentioned earlier in the literature, lifting the hip for the immediate adjustment of ground reaction forces can be used as a slippage recovery strategy \cite{park2001reflex}. However, the fact that higher stature is predictive of better balancing ability in dealing with slippage conditions is a new observation, applicable to both passive and active bipeds. 

\subsection{Push Recovery from Different Push Attitudes}

\begin{figure*}
 \centering
 \subfloat[]{\includegraphics[height=9 in-56.2 pt]{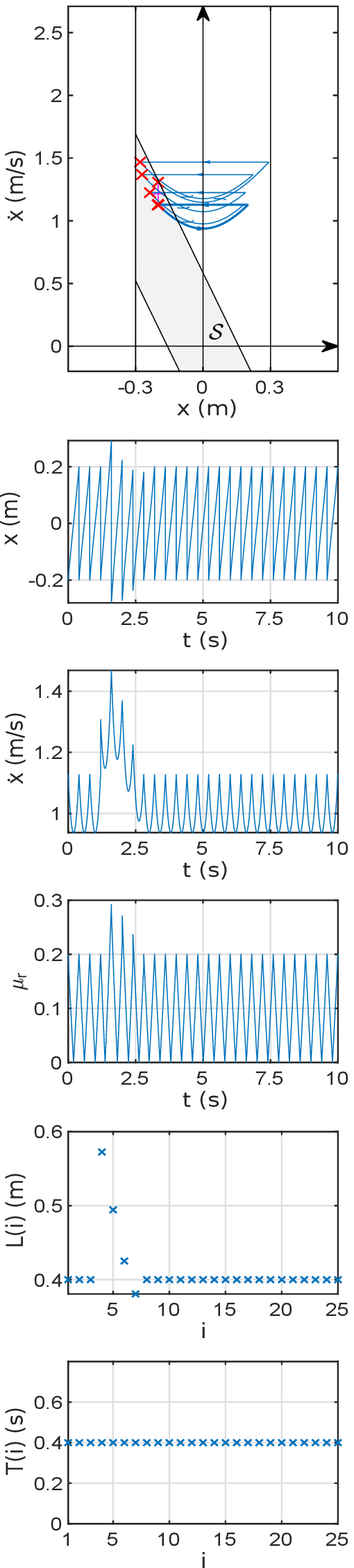}\label{fig:PushRecovery-FixedTime}}~
 \subfloat[]{\includegraphics[height=9 in-56.2 pt]{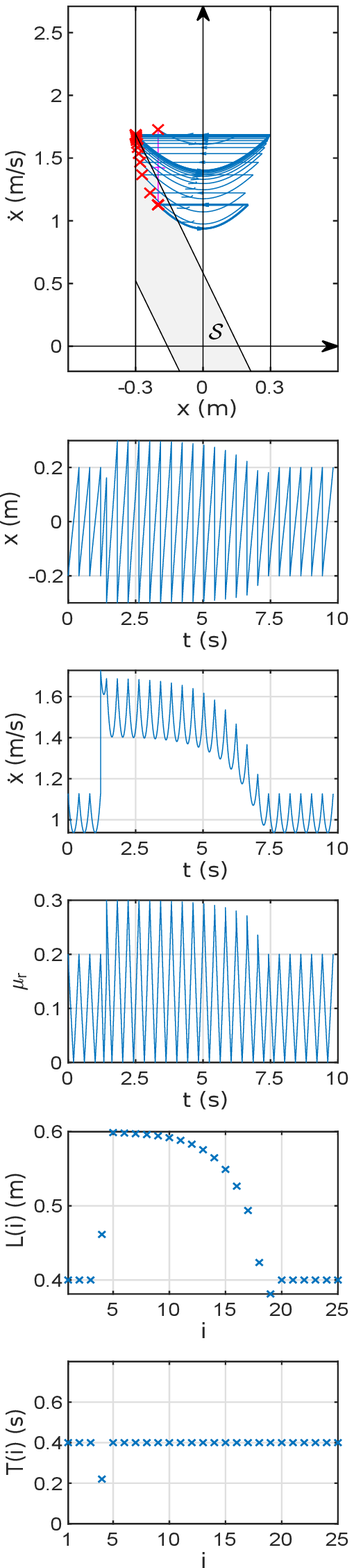}\label{fig:PushRecovery-FixedBorder}}~
 \subfloat[]{\includegraphics[height=9 in-56.2 pt]{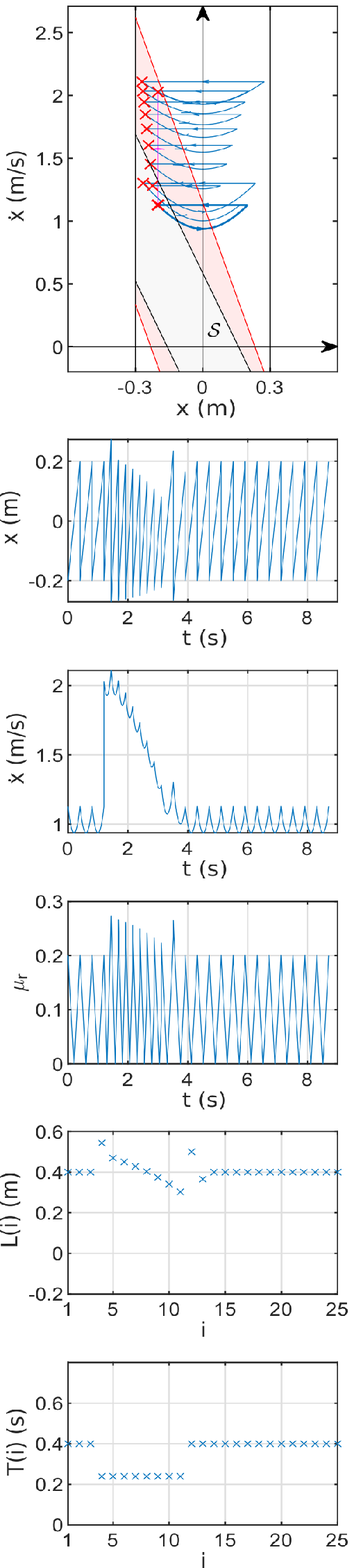}\label{fig:PushRecovery-MovingBorder}}
 \caption{Push recovery on a low friction surface ($\mu=0.3$) against different push magnitudes: \protect\subref{fig:PushRecovery-FixedTime} $\hat F=9$, \protect\subref{fig:PushRecovery-FixedBorder} $\hat F=30$, and \protect\subref{fig:PushRecovery-MovingBorder} $\hat F=45$ (all kg.m/s). The top subplots depict the phase portraits with limit cycles in thick plot, the initial states indicated in (red) `$\times$' markers, and the safe regions $\mathcal{S}$ colored (in gray or red).}
 \label{fig:PushRecovery}
 \end{figure*}

Consider the biped presented in part~\ref{subsec:Switching}, walking steadily on a low friction surface ($\mu=0.3$). At the beginning of step $i=4$, the biped is pushed from behind by an impulsive force applied horizontally at its CoM. The results of simulating such a scenario are depicted in Fig.~\ref{fig:PushRecovery} for three different pushing amplitudes, namely, $\hat F=9$, $\hat F=30$, and $\hat F=45$ (all kg.m/s). 
 
As depicted in the phase portraits, the controller is able to withstand the impacts of different magnitudes by reconverging the motion to the desired gait while keeping the initial states of each step in the safe region $\mathcal{S}$.
Accordingly (as shown in $\mu_r$ plots), the necessary friction coefficient $\mu_r$ at every instant is kept lower than $\mu$ that the surface can supply, in all three cases. 
However, the controller reacts differently to each case.

In the first case, Fig.~\ref{fig:PushRecovery-FixedTime}, where impact momentum isn't sufficient to throw the footstep initial state out of the safe region, the controller only needs to modify the step length without altering the step time for stabilizing the motion, as seen in the $L(i)$ and $T(i)$ plots. 

In the second case, Fig.~\ref{fig:PushRecovery-FixedBorder}, where the impact is more powerfully applied so that the footstep initial state is thrown out of the safe region, the controller changes both step-length and duration to withstand the impact. 
Recall that if the initial state is out of $\mathcal{S}$ but within $\mathcal{S}_0$, a subsequent slippage can be expected, preventable by step time adjustment.
Accordingly, as depicted in the $T(i)$ plot, the controller reacts by modifying the step time right after the impact event, thus preventing the state to exit the $\mathcal{S}_0$ border. Also, the after-impact step length is adjusted such that the next initial state returns into $\mathcal{S}$.
Hereafter, the motion convergence to the desired gait is achieved by exclusively changing the steps length.
Note that in this latter case, the fixed-border technique was applicable since the expelled initial state was still in $\mathcal{A}$, \eqref{eq:FixedBorderA}.

In the third case, Fig.~\ref{fig:PushRecovery-MovingBorder}, in which the impact is even more powerful, the initial state is found ousted deeply beyond the safe region, thus forcing the controller to use the moving-border technique in order to prevent slippage.
Recall that in the moving-border technique, the desired step time is altered to form a secondary safe region that can include the ousted state. As can be seen, the step time adaptation is continued for several consecutive steps after the instant of impact. Consequently, the motion convergence to the marching-in-place gait is followed by the step length adaptation until the initial state returns to the primary safe region. Subsequently, the step time is restored to its original desired value, and the motion returns to the desired gait. In all these stages, the initial states are kept within a primary or secondary safe region to prevent slipping.
Note that in this case, more step time adjustment attempts are needed in comparison to the second case. 

Compared to the state-of-art stabilizers such as \cite{de2012foot} and \cite{khadiv2020walking}, although footstep time adjustment has already been used for converging motion to the desired gait, it hasn't been employed for preventing slippage against external pushes so far. According to the present results, adjusting footstep time can even permit coping with significant push on slippery surfaces.

From the simulation results, one can infer that:
\begin{inparaenum}[(i)]	
\item
Step length adaptation in combination with step time adjustment proves to be more efficient in enhancing the ability to withstand strong disturbances without slippage even on low-friction surfaces;
\item 
In the case of strong disturbances, motion convergence to the desired gait is momentarily neglected for several steps to prevent slipping and reconsidered after reaching a safe state.
\end{inparaenum}

\subsection{6-DoF Planar Biped Push Recovery on a Low Friction Surface}

\begin{figure}
 \centering
 \subfloat[]{\includegraphics{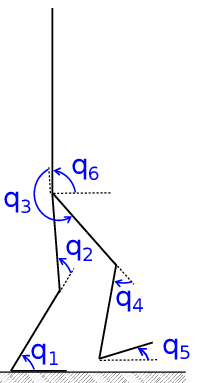}\label{fig:SimModel}}\\
 \subfloat[]{\includegraphics{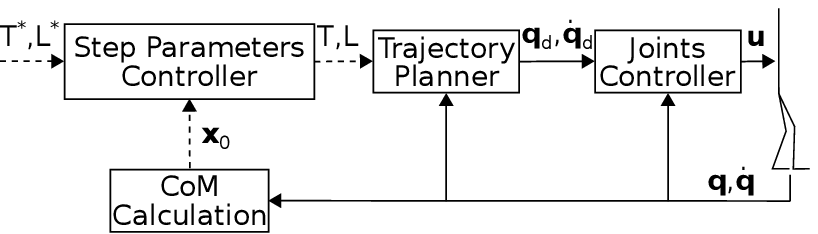}\label{fig:SimSchematic}}
 \caption{6-DoF planar biped simulation: \protect\subref{fig:SimModel} The simulation model; \protect\subref{fig:SimSchematic} The  control loop block diagram. The discrete signals are shown in dashed line.}
 \label{fig:SimModel-Schematic}
\end{figure}

In order to investigate the effectiveness of the proposed controller in more realistic conditions, a simulation on a six-degrees-of-freedom (6-DoF) planar model of a biped is performed. The simulation model and  schematics are illustrated in Fig.~\ref{fig:SimModel-Schematic}.
The biped model is considered as fully actuated, with three actuators for each leg.
The model parameters are set to be realistic, close to Nao's (a small size biped, commonly-used in labs due to its low price). The generalized coordinates are denoted by $\mathbf{q}\in\mathbb{R}^6$, where $\mathbf{q}=[q_1,q_2,q_3,q_4,q_5,q_6]^T$ is defined in Fig.~\ref{fig:SimModel}.
For the schematic block diagram depicted in  Fig.~\ref{fig:SimSchematic}, the governing equations of motion of the biped in SSP are derived by the Lagrangian method, and the DSP is modeled as an instantaneous event. The joints controller consists of the computed torque method, and the joints trajectory planner is developed based on the following optimization problem.
 
\begin{figure*}
 \centering
 \subfloat[]{\includegraphics[height=9 in-56.2 pt]{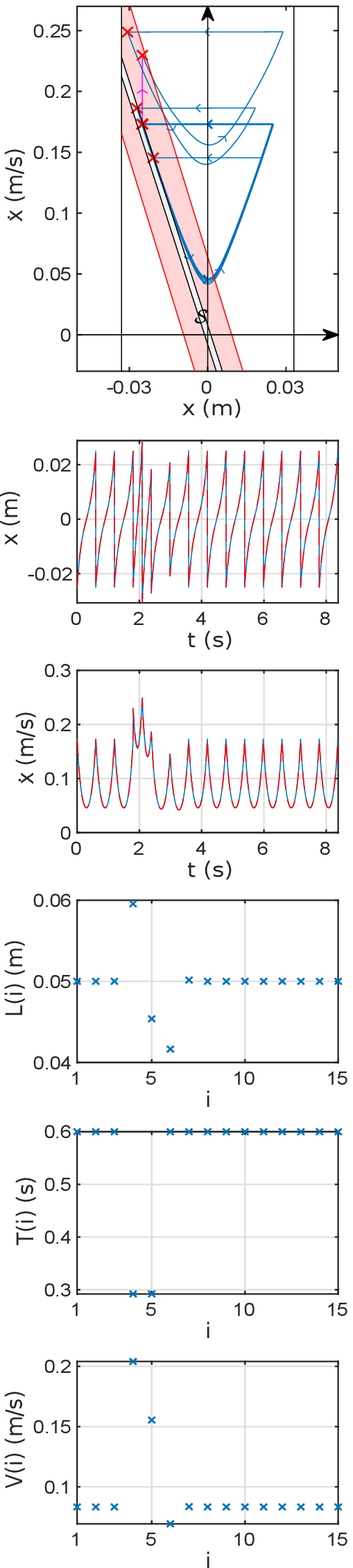}\label{fig:SixDoFCoM}}~
 \subfloat[]{\includegraphics[height=9 in-56.2 pt]{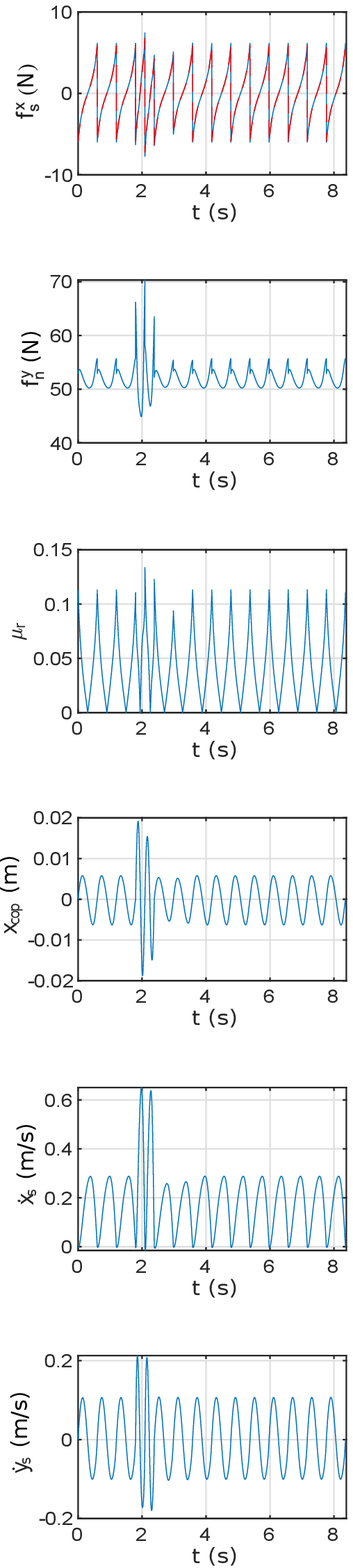}\label{fig:SixDoFFeasibility}}~
 \subfloat[]{\includegraphics[height=9 in-56.2 pt]{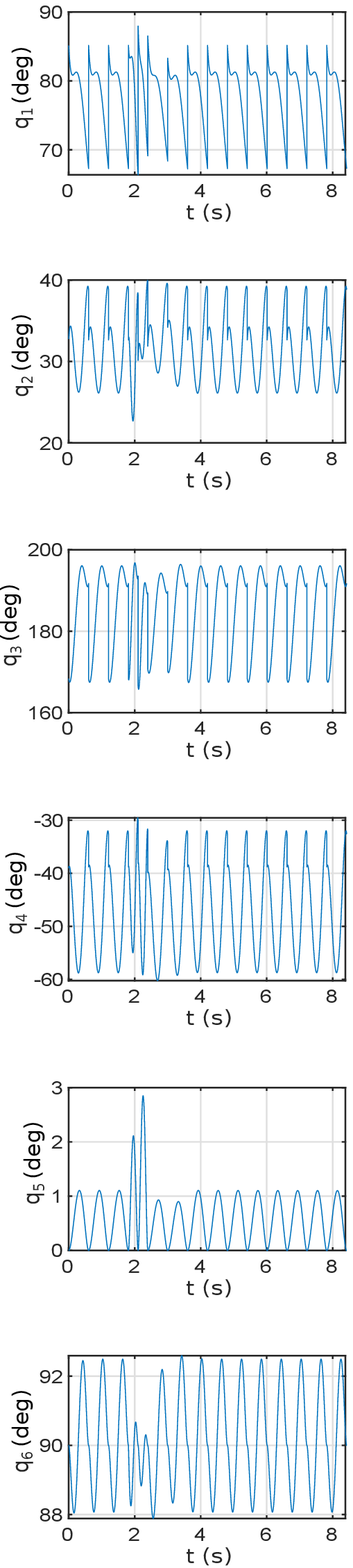}\label{fig:SixDoFqs}}
 \caption{6-DoF planar biped push-recovery on a low friction surface with $\mu=0.15$. In the phase portrait, limit cycles are shown thicker, the initial conditions indicated by (red) `$\times$' markers, and the primary and secondary safe regions $\mathcal{S}$ are shadowed (in red and gray, respectively). The (red) dashed lines are the trajectories corresponding to the linear dynamics.}
 \label{fig:SixDoFSimulation}
 \end{figure*}

By considering the following optimization index at every step,  
\begin{equation}\label{eq:OptimizationIndex}
J=\int_0^{T(i)}\left((\ddot x-\omega^2x)^2+\rho(y_s-y_s^d)^2\right) \mathrm{d}t,
\end{equation}
where $\rho\in\mathbb{R}^+$ is a weight coefficient, and $y_s,y_s^d\in\mathbb{R}^+$ are the vertical position of the swing leg (heel part) and its desired value, respectively. Minimizing such an index aims to ensure that the horizontal motion of the CoM is aimed to follow the linear dynamics \eqref{eq:SimplifiedModel} as near as possible, while enforcing that the swing leg heel follows a predefined trajectory. For fulfilling other requirements, the following constraints must apply:
\begin{enumerate}
\item
The position of the swing leg sole has to satisfy  $[x_s(T(i)),\allowbreak y_s(T(i))]=[L(i),0]$, where $x_s\in\mathbb{R}$ is its horizontal position so that the computed step length and time are met.
\item
The velocity of the swing leg sole has to satisfy  $[\dot x_s(T(i)),\allowbreak \dot y_s(T(i))]=[0,0]$ so that it touches the ground without bouncing back.
\item
The horizontal position and velocity of the CoM have to satisfy $[x(T(i)),\allowbreak\dot x(T(i))]=\mathbf{A}(T(i))\mathbf{x}_0^i$ so that the discrete dynamics \eqref{eq:MapEquation} applies.
\item
The vertical position and velocity of the CoM have to satisfy $[ y_{com}(T(i)),\allowbreak\dot {y}_{com}(T(i))]=[h,0]$ so that the CoM remains at a constant level, $h$. 
\item
The position of the CoP has to satisfy $[x_{cop}(0), \allowbreak x_{cop}(T(i))\allowbreak]\allowbreak=[0,0]$ so that CoP remains around the ankle.
\item
The angle and velocity of the swing leg sole have to satisfy $[q_5(T(i)),\allowbreak\dot q_5(T(i))]\allowbreak=[0,0]$ in order that the swing sole touches the ground horizontally.
\item
The angle and velocity of the torso have to satisfy \allowbreak $[q_6(T(i)),\dot q_6(T(i))]=[\pi/2,0]$ so that the torso remains nearly vertical.
\item
The knee angles have to satisfy $q_2(T(i)/2),q_2(T(i))\geq 0$ and  $q_4(T(i)/2),\allowbreak q_4(T(i))\leq 0$ so that their restrictions of movement are (likely) met.
\end{enumerate}
In the simulation, the above optimization is solved numerically using the nonlinear programming method. To this end, a fifth degree polynomial $\mathbf q=\mathbf q_0+\dot{\mathbf q}_0t+\mathbf c_1t^2+\mathbf c_2t^3+\mathbf c_3t^4+\mathbf c_4t^5$ is evoked as the desired joint trajectory, where $\mathbf q_0=\mathbf q(0)$ and $\mathbf c_i\in\mathbb R^{6\times1}$ are the unknown parameters derived from the optimization process. 
 The sagittal plane scene presents the robot walking with the step length $L^*=0.05$ m,  the step time $T^*=0.6$ s, and the initial state ${\mathbf{x}_0^*=[-0.025\text{ m},0.173\text{ m/s}]^\top}$. The floor is considered of low-friction with $\mu=0.15$.
At the beginning of step $i=4$, the biped is being pushed from behind by an impact force $\hat{F}=0.3$ kg.m/s, applied horizontally to its torso midpoint. 
In the simulation, the impact is implemented as a jump in the joints' velocities, obtained by 
\begin{equation}
\dot{\mathbf q}^+=\mathbf M^{-1}\hat{\mathbf Q}+\dot{\mathbf q}^-
\end{equation}
\cite{baruh1999analytical}, where superscripts $\cdot^-$ and $\cdot^+$ denote the  before/after impact instances,  $\mathbf M\in\mathbb{R}^{6\times6}$ is the inertia matrix, and $\hat{\mathbf Q}\in\mathbb{R}^6$ is the generalized impact vector. 
The simulation results are depicted in Fig.~\ref{fig:SixDoFSimulation}.
 
As depicted in the phase portrait, the safe region $\mathcal{S}$ is so narrow that the next initial state is thrown out of it by the impact. As a reaction to this event, the controller opted to momentarily change the step-time duration for several consecutive steps to form a secondary safe region that includes the outed state. By adapting the step-length, a reestablishment maneuver consists of temporarily conducting the state toward the origin (equivalent to a marching-in-place gait). After the initial state is brought back to the primary safe region, the step-time is restored to its initial desired value, thus reinstating the desired gait within the primary safe region. The trend of step-length and step-time variations due to this impact event are illustrated in $L(i)$ and $T(i)$ plots, respectively.

The footstep velocity, denoted by $V(i)$, is illustrated in Fig.~\ref{fig:SixDoFCoM}. As shown, the controller first increases the speed after the impact, then decreases it until regaining the desired speed. 
Similar behavior can be observed in human response to powerful pushes, from which one can infer that: 
\emph{Increasing the walking speed as a reaction against powerful imposed disturbances helps to escape from slippage.}

The simulation can be validated based on the results depicted in Fig.~\ref{fig:SixDoFFeasibility} as follows:
\begin{itemize}[$\bullet$]
\item \emph{The support leg sole will not disconnect from the ground}.
As can be seen in the $f_n^y$ plot, the normal force $f_n^y$ remains around a positive constant value due to the trajectory planner constraint of keeping the CoM at a constant height. 
\item \emph{The support leg will not slip}.
As the $\mu_r$ plot shows, the necessary friction coefficient $\mu_r$ at every instant remained lower than the value $\mu$ that the surface can supply.
\item \emph{The motion is executable on the actual biped model}.
The plot of the CoP position, $x_{cop}$, indicates that the designed motion is feasible if the biped has a minimal support sole of 4 cm length, with 2 cm behind the ankle; A condition fulfilled by the biped considered for this purpose.
\item \emph{There is no impact at each step}.
The $\dot x_s$ and $\dot y_s$ plots show that the swing leg sole touches the ground smoothly, at zero velocity.
\end{itemize}

\begin{figure}
\centering
\includegraphics{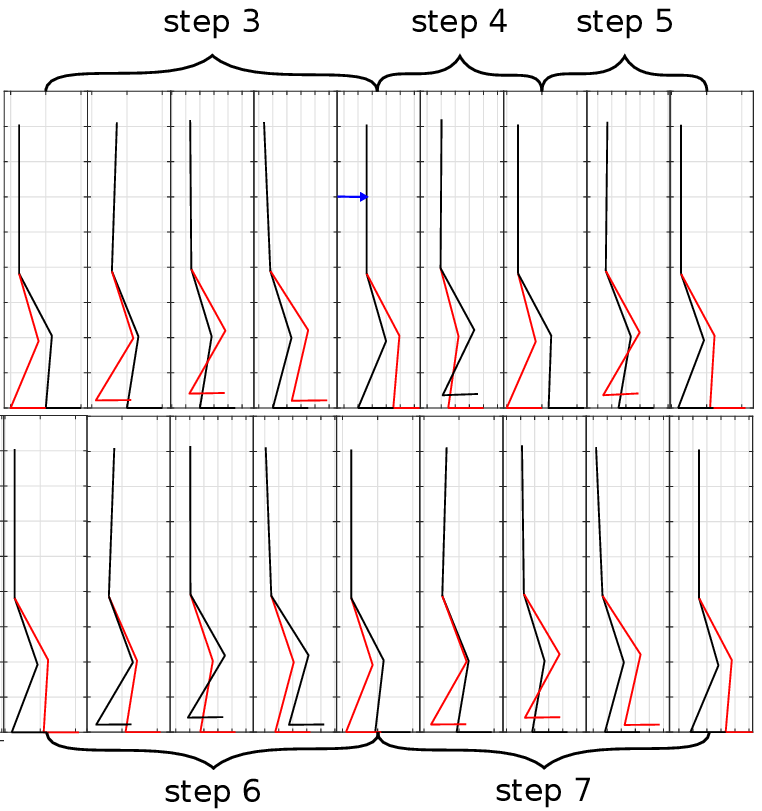}
\caption{A few frames taken at regular intervals (0.15 s) of the biped's motion including the step before the impact and several steps after it. Each step has been identified by braces (of different lengths). The impact force is shown by a (blue) horizontal vector applied to the torso at the beginning of step 4. Clip in the accompanying video\footnotemark}
\label{fig:Animation}
\end{figure}

The joints trajectories are depicted in Fig.~\ref{fig:SixDoFqs}. As can be seen, they are bounded and demonstrate repetitive patterns. The knees angles $q_2$ and $q_4$ satisfy their physical restrictions, the swing sole remains horizontal as the $q_5$ plot shows, and the torso remains nearly vertical as seen in the $q_6$ plot. The resulting motion is illustrated in Fig.~\ref{fig:Animation} by some snapshots taken at regular intervals through the animation sequence.

\setcounter{footnote}{-1}
Results demonstrated that the step length-time controller strategy is successful in stabilizing the biped walking motion on low friction surfaces, reacting just as humans would do.
However, for the implementation of the controller on an active (actuated) biped, a proper trajectory planner is necessary to generate the required motion, constraining it to conform with the linear dynamics of \eqref{eq:MapEquation} and to satisfy its assumptions as best as possible. 
On the other hand, this trajectory planner reduces the flexibility of the biped to correctly react to impacts, limiting the range of tolerable disturbances. Moreover, as real-world applications necessitate fast reactions, a significant concern subsists with regard to the computational time necessary for the optimization trend of the trajectory planner. 
Thus, one can infer that a compromise should be reached between satisfying the dynamic model based on which these particular step time and step length controllers were obtained, and relaxing these conditions for the purpose of real-time implementation and flexibility. 
Note, though we implemented the step controller on an fully actuated biped in the simulation, it is also implementable (mostly with a lower degree trajectory planner) on semi-active bipeds with control on the step length and time. 


\section{Discussion}

In this section, first, a brief discussion on the effectiveness of the method for extending its usage in 3-dimensional space (3D) is conducted. Then, the advantages of the proposed controller over those proposed in the literature are summed up. Finally, the limitations of the method are investigated at the end.

\subsection{Method Effectiveness in 3D}

\begin{figure}
\centering
\includegraphics{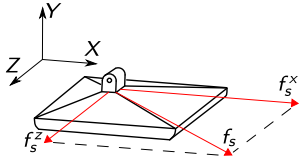}
\caption{Horizontal reaction forces applied to the foot from the ground}
\label{fig:FrictionComponents}
\end{figure}

\begin{figure}
\centering
\includegraphics{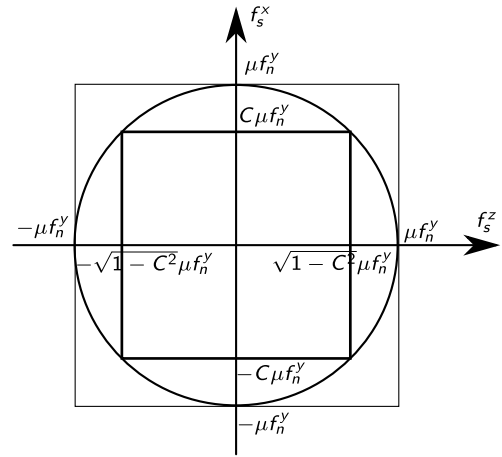}
\caption{Schematic of friction circle, circumscribed square, and inscribed rectangle}
\label{fig:FrictionCircle}
\end{figure}

To investigate the plausibility of extending the proposed method to the more realistic 3D walking situation, first consider the reaction forces applied to the foot from the ground as depicted in Fig.~\ref{fig:FrictionComponents}. 
As shown in the figure, in contrast to the 2D model, an extra lateral force component $\overrightarrow{f_s^z}$ arises at the foot sole, mainly due to the lateral oscillating motion of the biped during walking.
Accordingly, the general non-slipping condition can be expressed as $|f_s|<\mu f_n^y$, where $\overrightarrow{f_s}=\overrightarrow{f_s^x}+\overrightarrow{f_s^z}$ is the resultant horizontal reaction force.
The non-slippage condition in terms of the reaction forces components can thus be expressed as
\begin{equation}\label{eq:3DNonSlippageCondition}
(f_s^x)^2+(f_s^z)^2<(\mu f_n^y)^2,
\end{equation}
which is geometrically interpreted as follows: The resultant of longitudinal and lateral components should lie within the boundary of the friction circle (Fig.~\ref{fig:FrictionCircle}) to prevent slippage. 
\footnotetext{\url{https://youtu.be/BWzUgHGdl3I}}
Considering the friction circle, one can see the following obvious but critical points:
\begin{itemize}[$\bullet$]
\item 
Satisfying the two conditions $|f_s^x|<\mu f_n^y$ and $|f_s^z|<\mu f_n^y$ independently of each other is not sufficient to prevent slippage. Indeed, those conditions identify to a  circumscribed square of the friction circle (Fig.~\ref{fig:FrictionCircle}), lying outside of the safe region.
\item
Indeed the two reaction force components cannot be independent; As the magnitude of one grows, the other should decay so that their resultant remains within the friction circle.
\end{itemize}
According to the above discussion, our 2D model-based controller may cause the biped to slip in a real-world 3D walking task since the whole friction capacity of the surface is entirely assigned to the longitudinal motion in the algorithm. 
In this regard, there may not remain sufficient friction capacity to maintain the lateral swinging motion, causing the biped to slip.
However, one can deal with this problem by re-scaling the upper bound of the longitudinal non-slippage condition, \ie, $|f_s^x|<C\mu f_n^y$ where $0<C<1$.
Consequently, there will always remain a sufficient friction capacity for bearing the lateral component bounded to $|f_s^z|<\sqrt{1-C^2}\,\mu f_n^y$, according to \eqref{eq:3DNonSlippageCondition}.   
Therefore, the present method for stabilizing the biped in the sagittal plane is still effective in real-world 3D walking if the friction coefficient is reduced, \ie, $\mu$ is modified by a coefficient $0<C<1$ in the formulas. Subsequently, this same algorithm can be employed for setting the lateral step size (balancing motion in the sagittal plane) by considering a friction coefficient equal to   $\sqrt{1-C^2}\,\mu$, which alongside the longitudinal non-slippage condition, can guarantee 3D walking without slip.
In this manner, the surface friction capacity is distributed between the longitudinal and lateral motions. By selecting an appropriate value for $C$, depending on the situation, the lateral/longitudinal step-length and step-time controllers will adjust accordingly to ensure non-slippage.

\subsection{Comparison with the State of the Art}

Footstep adjustment has been widely  used in the literature for preserving gait stability in confronting external disturbances \cite{yu2018disturbance,de2012foot,khadiv2020walking,faraji2019bipedal}.
The main limitation of such stabilizers is that the surface friction limitation is not included in their design, \ie, there is no distinction between slipperiness of surfaces, which may induce a disproportional reaction, conducting to slippage, and a complete loss of stability.
Those stabilizers are unreliable, especially on slippery surfaces whereas, by respecting the surface capacity, our approach results in push recovery capability on slippery surfaces.
On the other hand, footstep duration adjustment has a completely different and more specific aim in our work compared to the literature.
In the state-of-art stabilizers found in the literature, both footstep duration and length adjustment are spent on  converging the gait to the desired one \cite{de2012foot,khadiv2020walking}. 
In contrast, in the present work, the footstep time adjustment part is exclusively charged for preventing (subsequent) slippage, thus providing a firm basis for converging to the desired gait via the footstep length adjustment part.
As the results show, taking charge of this single task by footstep duration adjustment enhances the capability of coping with disturbances on slippery surfaces and results in human-like reactions in confronting significant disturbances.
Moreover, this controller has an analytical form, consisting of some simple closed-form solutions for computing the control inputs, in contrast to the non-deterministic optimization methods existing in the literature \cite{khadiv2020walking}, which nevertheless also employed simplified dynamics models of the biped.
On the other side, the main focus of studies on biped walking on slippery surfaces is gait generation in the absence  of  external disturbances \cite{brandao2016footstep,feng20133d,kajita2004biped,ma2019dynamic,park2001reflex,chen2021robust}. 
The proposed stabilizer here complements those planners by stabilizing generated gaits in the confrontation of external disturbances.

\subsection{Limitations}

The proposed controller in this paper is constructed based on a simplified model of the biped, restricting the center of mass to move horizontally and neglecting any fluctuation in the overall angular momentum. For implementing those abstract control rules on an active biped, although constraining the joints trajectory planner to conform with the model and its assumptions reduces the modeling mismatch error, it reduces the flexibility of the biped in confronting more or less powerful disturbances.

The proposed controller also relies on the knowledge of the lower/upper bound of the surface friction coefficient. While the estimation might be difficult in practice, it is believed that material classification from images and tabled coefficient of frictions is a feasible approach to the problem \cite{brandao2016footstep}. In addition, the controller can still be applied when some uncertainty in the estimation is considered, \eg, by adding a margin to the no-slippage constraint depending on the estimated uncertainty.


\section{Conclusion}

In this paper, a stabilizer for biped sagittal motion on slippery surfaces has been proposed. 
By introducing some assumptions, a reduced dynamic model with respect to the CoM variables of the biped is derived, leading to a discrete map relating the initial state to other consecutive steps. Via this model, a cyclic gait of desired speed can be adjusted through footstep-length and time parameters serving as control inputs. The controller is a discrete event-based one, which sets the footstep length and duration at the beginning of each step.
Defining the safe region as the set of footstep initial states that ensures the non-slipping conditions during the step, the step length is determined such that: 
\begin{inparaenum}[(i)]
\item the footstep initial state converges to its desired value, while
\item the next footstep initial state is kept within the no-slippage safe region.
\end{inparaenum}
Using LaSalle's theorem, it is proven that the designed step-length controller provides asymptotical convergence of the motion to the cyclic gaits while respecting non-slipping conditions.

On the other hand, in the cases where the footstep initial state is found ousted from the safe region, \eg\ by the intervention of a powerful pushing force, the footstep duration is reduced by the step-time controller to prevent subsequent slippage. 
Based on the ousted distance from the safe region, two time-adjustment techniques which extend the safe region to include the ousted state can be employed. In the first approach, only the duration of the eminent-slipping step is adapted momentarily to re-adhere the state to the safe region. Whereas, in the other method, the desired step time is altered for several consecutive steps afterward, displacing the borders and shaping another safe region.
 
As the designed stabilizer uses elementary math operations, its computation cost is low, so it is adapted for real-time implementation on an actual system, even with the modest computational hardware. 

The soundness of the proposed controller has been tested through several numerical simulations with various scenarios, including performing gait switching maneuvers on different grounds and push recovery on a low-friction surface. 
The results show that the designed stabilizer has a good performance in maintaining the stability of motion on slippery surfaces in confronting significant disturbances and exhibits responses similar to human reactions.
In correlation with human gait experience, the results also reveal some general biomechanics aspects of biped walking which may be beneficial in developing bio-inspired control algorithms for bipeds, summarized as the following:

\begin{itemize}[$\bullet$]
\item
Biped CoM height has a direct effect on reducing the risk of slipping. 
This reveals that ''\emph{Taller persons are less prone to slippage compared to shorter ones when executing similar gaits}''.
Consequently, changing CoM height can be employed as an alternative for reducing the risk of slipping, both in the designing stage (especially for passive walkers) and in the gait generation/adaptation phase.
\item
Footstep duration adjustment can play an exclusive role in slippage prevention in confronting external pushes. 
It is observed that the reaction of the controller for escaping from slippage when exposed to powerful disturbances is to increase the walking speed through decreasing footstep duration. 
Similar behavior can be observed in human response to powerful pushes, from which one can infer that this reaction is a  dynamical adaptation to escape slippage. 

\end{itemize}

For future work, the feasibility of performing a real-time implementation of the developed method on a humanoid robot should be evaluated, considering that a low-level joints trajectory planner has to be developed in accordance with the high-level controller, compromising between satisfying the assumptions of the dynamic model, the real-time implementation and the flexibility of the movements.

\section*{Declaration of Competing Interest}
The authors declare that they have no known competing financial interests or personal relationships that could have appeared to influence the work reported in this paper.

\appendix
\begin{appendices}

\section{Proof of Proposition~\ref{th:SafeRange}}
\label{prf:Existance}

As we discussed in Section \ref{sec:frictionAnalysis}, if $\mathbf{x}_0^i \in \mathcal{S}$ and $L(i) \in \left(L_l^{s}(i),L_u^{s}(i)\right)$, then $\mathbf{x}_0^{i+1} \in \mathcal{S}$.
Considering \eqref{eq:SextSnxt}, one can conclude that $\mathcal{S} \subset \mathcal{S}_T$.
Thus, for $\mathbf{x}_0^{i+1}\in \mathcal{S}$, we have $\mathbf{x}_0^{i+1} \in \mathcal{S}_T$, and according to \eqref{eq:STComputation},
\begin{equation}
\frac{-A_{12}\dot{x}_0^{i+1}-\mu h}{A_{11}}<x_0^{i+1}<\frac{-A_{12}\dot{x}_0^{i+1}+\mu h}{A_{11}}.
\end{equation}
Using the above inequality and the difference equation~\eqref{eq:MapEquation}, one can conclude that
\begin{equation}\label{eq:LemmaP1}
\frac{\dot{x}_0^{i+1}-A_{21}\mu h}{A_{11}}<\dot{x}_0^{i+2}<\frac{\dot{x}_0^{i+1}+A_{21}\mu h}{A_{11}}.
\end{equation}
Moreover, Fig.~\ref{fig:S} shows that for $\mathbf{x}_0^{i+1}\in \mathcal{S}$,
\begin{equation}
    |\dot{x}_0^{i+1}|<\frac{A_{11}+1}{A_{12}}\mu h.
\end{equation}
The above inequality and Property~\ref{prop:AMatrixProperties} result that
\begin{equation}\label{eq:LemmaP2}
\begin{dcases}
\frac{\dot{x}_0^{i+1}-A_{21}\mu h}{A_{11}}>-\frac{A_{11}+1}{A_{12}}\mu h,\\
\frac{\dot{x}_0^{i+1}+A_{21}\mu h}{A_{11}} < \frac{A_{11}+1}{A_{12}}\mu h,
\end{dcases}
\end{equation}
According to \eqref{eq:LemmaP1} and \eqref{eq:LemmaP2}, one concludes that $|\dot{x}_0^{i+2}|<\frac{A_{11}+1}{A_{12}}\mu h$.

On the other hand, the safe range $\left(L_l^{s}(i+1),L_u^{s}(i+1)\right)$ exists iff $L_l^{s}(i+1)<L_u^{s}(i+1)$.
According to \eqref{eq:LlSafe} and \eqref{eq:LuSafe}, we have $L_l^{s}(i+1) < L_u^{s}(i+1)$ iff the following four conditions hold,
\begin{enumerate}
    \item $ L_l^0(i+1) < L_u^0(i+1) $,
    
        Considering \eqref{eq:Ll0} and \eqref{eq:Lu0}, ${L_l^0(i+1) < L_u^0(i+1)}$ is always satisfied.
        
    \item $ L_l^0(i+1) < L_u^T(i+1) $,
    
        According to \eqref{eq:Ll0} and \eqref{eq:LuT}, the inequality $L_l^0(i+1) < L_u^T(i+1)$ can be rewritten as
        \begin{equation}
            \left(A_{11}^2-1\right) x_0^{i+1} + A_{11}A_{12}\dot{x}_0^{i+1} > -\left(A_{11}+1\right) \mu h.
        \end{equation}
        Using the difference equation~\eqref{eq:DSPdynamics} and Property~\ref{prop:AMatrixProperties}, the above inequality is simplified as
        \begin{equation}
            \dot{x}_0^{i+2} > -\frac{A_{11}+1}{A_{12}}\mu h.
        \end{equation}
        
    \item $L_l^T(i+1) < L_u^0(i+1) $,
    
        According to \eqref{eq:LlT} and \eqref{eq:LlT}, the inequality $L_l^T(i+1) < L_u^0(i+1)$ can be rewritten as
        \begin{equation}
            \left(A_{11}^2-1\right) x_0^{i+1} + A_{11}A_{12}\dot{x}_0^{i+1} < \left(A_{11}+1\right) \mu h.
        \end{equation}
        Using the difference equation~\eqref{eq:DSPdynamics} and Property~\ref{prop:AMatrixProperties}, the above inequality is simplified as
        \begin{equation}
            \dot{x}_0^{i+2} < \frac{A_{11}+1}{A_{12}}\mu h.
        \end{equation}
        
    \item $L_l^T(i+1) < L_u^T(i+1)$.
    
        Considering \eqref{eq:LlT} and \eqref{eq:LuT}, one concludes that $L_l^T(i+1) < L_u^T(i+1)$ is always satisfied.
\end{enumerate}

The above explanation results that $\left(L_l^{s}(i+1),L_u^{s}(i+1)\right) \neq \varnothing$ iff
\begin{equation}
\label{eq:PropMain}
    |\dot{x}_0^{i+2}| < \frac{A_{11}+1}{A_{12}}\mu h.
\end{equation}

In conclusion, for $L(i) \in \left(L_l^s(i),L_u^s(i)\right)$ and $\mathbf{x}_0^i \in \mathcal{S}$, we have $|\dot{x}_0^{i+2}| < \frac{A_{11}+1}{A_{12}}\mu h$, and hence $L_l^{s}(i+1) < L_u^{s}(i+1)$, thus $\left(L_l^s(i+1),L_u^s(i+1)\right)$ exists.

\end{appendices}
\begin{appendices}

\section{Proof of Proposition~\ref{th:ConvergenceRange}}
\label{prf:Convergence}

Define a continuous scalar function as
\begin{equation}\label{eq:lyapunoveFcn}
{V(\Delta \mathbf{x}_0)}={\left(A_{21}\Delta {x}_0+A_{22}\Delta \dot{x}_0\right)^2}.
\end{equation}
Considering \eqref{eq:ErrorDynamics}, ${\Delta V(\Delta \mathbf{x}_0^i)=V(\Delta \mathbf{x}_0^{i+1})-V(\Delta \mathbf{x}_0^i)}$ is obtained as
\begin{equation}
\Delta V=A_{21}^2\left(\Delta L(i)-\Delta L_1(i)\right)\left(\Delta L(i)-\Delta L_2(i)\right). 
\end{equation}
If $\Delta L(i) \in \left( \min \{\Delta L_1(i), \Delta L_2(i)\},\max\{\Delta L_1(i), \Delta L_2(i)\} \right)$ then $\Delta V\leq 0$ (where equality holds when $\Delta L_1(i)=\Delta L_2(i)$ and  the interval has shrunk to a single value).
Therefore, if \eqref{eq:convergenceRange} is satisfied, $V$ is a liapunov function\footnote
{
Consider a discrete-time dynamical system as
\begin{equation*}
\mathbf{x}(k+1)=f(\mathbf{x}(k)).
\end{equation*}
For the above system, the scalar value function $V(\mathbf{x})$ is called a Liapunov function on a set like $\mathcal{G}$ if
\begin{inparaenum}[(i)]
\item $V$ is continuous, and 
\item $\Delta V(\mathbf{x}) = V(f(\mathbf{x}))-V(\mathbf{x})\leq 0$ for all $x\in \mathcal{G}$ \cite{la1976stability}.
\end{inparaenum}\vspace{5pt}
}
of \eqref{eq:ErrorDynamics}.

On the other hand, define $\mathcal{E}$ as
\begin{equation*}
\mathcal{E}=\{\Delta \mathbf{x}\mid\Delta V=0\}.
\end{equation*}
If $\Delta L(i)$ satisfies \eqref{eq:convergenceRange}, one can conclude that
\begin{equation}
\mathcal{E}=\left\{\,\Delta \mathbf{x}_0\mid A_{21}\Delta {x}_0+A_{22}\Delta \dot{x}_0=0\,\right\}.
\end{equation}
For every $\Delta \mathbf{x}_0^i\in \mathcal{E}$, since $\Delta \dot{x}_0^{i+1}=A_{21}\Delta {x}_0^i+A_{22}\Delta \dot{x}_0^i=0$,  we have $\Delta \mathbf{x}_0^{i+1}\in \mathcal{E}$ iff $\Delta {x}_0^{i+1}=0$.
Hence, the largest invariant set of $\mathcal{E}$, called $\mathcal{M}$, contains only one point, namely $\Delta{\mathbf{x}}_0=[0,0]^\top$.

Moreover, since $V$ is a non-increasing function, and $V(\Delta \mathbf{x}_0^i)=(\Delta \dot{x}_0^{i+1})^2$, one concludes that
$\Delta \dot{x}_0$ is also non-increasing.
Since $V$ and $\Delta \dot{x}_0$ are bounded, \eqref{eq:lyapunoveFcn} results that $\Delta x_0$ is also bounded.
In conclusion, every solution of \eqref{eq:ErrorDynamics} is bounded.

According to the above explanations, the LaSalle's theorem\footnote
{
\textbf{Theorem} (LaSalle's theorem)\textbf{.} 
\textit{
Consider a discrete-time dynamical system as
\begin{equation*}
\mathbf{x}(k+1)=f(\mathbf{x}(k)).
\end{equation*}
Let $\mathcal{G}$ be a bounded open positively invariant set.
If 
\begin{inparaenum}[(i)]
\item $V(\mathbf{x})$ is a Liapunov function of the above system on $\mathcal{G}$,
\item $\mathcal{M} \subset \mathcal{G}$ is the largest invariant set in $\mathcal{E}$, and
\item $V(\mathbf{x})$ is constant on $\mathcal{M}$,\label{th:LaSall3rdCond}
\end{inparaenum}
then $\mathcal{M}$ is asymptotically stable (globally asymptotically stable relative to $\mathcal{G}$) \cite{la1976stability}.
}\vspace{5pt}

The condition \eqref{th:LaSall3rdCond}  is automatically satisfied if $\mathcal{M}$ is a single point.
}
results that $\mathcal{M}$ is globally asymptotically stable, \ie\ $\Delta \mathbf{x}_0=\mathbf{0}$ is globally asymptotically stable.

\end{appendices}
\begin{appendices}

\section{Proof of Proposition~\ref{th:SafeConvRangeIntersection}}
\label{prf:Intersection}

We show that  the convergence range bound ${\Delta L_2(i)+L^*}$ is always in the safe range $\left(L_l^s(i),L_u^s(i)\right)$.  According to \eqref{eq:Ll0}, \eqref{eq:ConvRoot2}, and  Property~\ref{prop:AMatrixProperties}, one can show that
\begin{equation}
\begin{aligned}
L_l^0(i)<\Delta L_2(i)+L^*&\iff-\mu h<\frac{\left(A_{22}-1\right)\Delta \dot{x}_0^{i+1}}{A_{21}},\\
&\iff-\frac{A_{11}+1}{A_{12}}\mu h<\Delta \dot{x}_0^{i+1}.
\end{aligned}
\end{equation}
Doing the same as the above for other elements of $L_l^s(i)$ and $L_u^s(i)$ leads to
\begin{equation}
L_l^s(i)<\Delta L_2(i)+L^*<L_u^s(i)\iff|\Delta \dot{x}_0^{i+1}|<\frac{A_{11}+1}{A_{12}}\mu h.
\end{equation}
According to \eqref{eq:PropMain}, the above holds iff the safe range exists. Indeed, if the safe range exists then it has an intersection with the convergence range, and the theorem is thus proved.

\end{appendices}

\bibliographystyle{spmpsci}    
\bibliography{References} 

\end{document}